%% file: main.tex
\documentclass[10pt,twocolumn,letterpaper]{article}

\usepackage[pagenumbers]{cvpr}      %

\input{preamble}
\definecolor{cvprblue}{rgb}{0.21,0.49,0.74}
\usepackage[pagebackref,breaklinks,colorlinks,allcolors=cvprblue]{hyperref}
\usepackage{array}
\usepackage{pifont}
\usepackage{amsthm}
\usepackage{fontawesome5}
\usepackage[table]{xcolor}
\input{macros}

\def\confName{CVPR}

\title{\vspace*{-27pt}Teach with Blinks: Breaking Linguistic Momentum in On-Policy Distillation for Fine-Grained Multimodal Perception\par\vspace*{-8pt}}

\author{Tangzheng~Lian$^{1}$ \quad Hang~Zhang$^{2}$ \quad Guanyu~Hu$^{3}$ \quad Renjiu~Hu$^{2}$ \\
Zhensong~Zhang$^{5}$ \quad Jiankang~Deng$^{4}$ \quad Jifei~Song$^{5}$ \\
$^{1}$King's College London \quad $^{2}$Cornell University \quad $^{3}$Aalto University \\
$^{4}$Imperial College London \quad $^{5}$Huawei Noah's Ark Lab\\[-7pt]}

\begin{document}
\maketitle
\input{sec/0_abstract}
\input{sec/1_intro}
\input{sec/3_analysis}
\input{sec/4_method}
\input{sec/5_experiments}

\input{sec/6_conclusion}
\par
{
    \small
    \bibliographystyle{ieeenat_fullname}
    \bibliography{main}
}

\input{sec/X_suppl}

\end{document}

%% file: macros.tex
\usepackage{multirow}
\usepackage{placeins}  %
\usepackage{makecell}
\usepackage{bm}
\usepackage{algorithm}
\usepackage{algpseudocode}
\usepackage{tikz}
\usetikzlibrary{arrows.meta,positioning,calc,fit,backgrounds,decorations.pathreplacing}
\usepackage{pgfplots}
\pgfplotsset{compat=1.18}
\usepgfplotslibrary{groupplots,fillbetween}
\pgfplotsset{every axis/.append style={/pgf/number format/assume math mode=true}}

\makeatletter
\newwrite\supp@aux
\AtBeginDocument{\if@filesw\immediate\openout\supp@aux=\jobname-supp.aux\fi}
\def\supp@sortcites#1{\let\supp@mainaux\@auxout\let\@auxout\supp@aux
  \supp@origsortcites{#1}\let\@auxout\supp@mainaux}
\AtBeginDocument{\let\supp@origsortcites\NAT@sort@cites}
\newcommand{\citesupp}[1]{{\let\NAT@sort@cites\supp@sortcites\cite{#1}}}
\newcommand{\citealpsupp}[1]{{\let\NAT@sort@cites\supp@sortcites\citealp{#1}}}
\newcommand{\bibliographystylesupp}[1]{%
  \if@filesw\immediate\write\supp@aux{\string\bibstyle{#1}}\fi}
\newcommand{\bibliographysupp}[1]{%
  \if@filesw\immediate\write\supp@aux{\string\bibdata{#1}}\fi
  \begingroup
    \edef\supp@start{\the\c@NAT@ctr}%
    \let\supp@thebibliography\thebibliography
    \def\thebibliography##1{%
      \supp@thebibliography{\number\numexpr##1+\supp@start\relax}%
      \global\c@NAT@ctr=\supp@start\relax}%
    \input{sec/X_suppl_refs}  \endgroup}
\makeatother

\newcommand{\method}{BlinkOPD\xspace}     %
\newcommand{\methodfig}{BlinkOPD}
\newcommand{\methodlong}{Blink On-Policy Distillation\xspace}

\DeclareMathOperator*{\argmax}{arg\,max}
\DeclareMathOperator{\softmax}{softmax}

\newcommand{\pT}{p^{+}}                 %
\newcommand{\pC}{p^{\varnothing}}       %
\newcommand{\pR}{p^{\mathrm{ref}}}      %

\newcommand{\cmark}{\ding{51}}
\newcommand{\xmark}{\ding{55}}
\definecolor{oursrow}{RGB}{232,240,252}
\newcommand{\gain}[1]{{\color{ForestGreen}#1}}
\newcommand{\drop}[1]{{\color{BrickRed}#1}}
\newcommand{\rep}[1]{{\color{black!55}#1}}
\newcommand{\grouprow}[2]{\rowcolor{gray!15}\multicolumn{#1}{c}{\textit{#2}}\\}

\definecolor{cOurs}{RGB}{214,39,40}
\definecolor{cAha}{RGB}{31,119,180}
\definecolor{cVAD}{RGB}{118,42,184}
\definecolor{cCVOPD}{RGB}{44,160,44}
\definecolor{cOPDV}{RGB}{255,127,14}
\definecolor{cVOPD}{RGB}{127,127,127}
\definecolor{cGray}{RGB}{90,90,90}

%% file: sec/X_suppl_refs.tex
\begin{thebibliography}{10}
\providecommand{\natexlab}[1]{#1}
\providecommand{\url}[1]{\texttt{#1}}
\expandafter\ifx\csname urlstyle\endcsname\relax
  \providecommand{\doi}[1]{doi: #1}\else
  \providecommand{\doi}{doi: \begingroup \urlstyle{rm}\Url}\fi

\bibitem[Akhondzadeh et~al.(2026)Akhondzadeh, Lingam, Tejaswi, Ekbote,
  Sanghavi, and Bojchevski]{akhondzadeh2026rgopd}
Mohammad~Sadegh Akhondzadeh, Vijay Lingam, Atula Tejaswi, Chanakya Ekbote,
  Sujay Sanghavi, and Aleksandar Bojchevski.
\newblock Reward-gated on-policy distillation.
\newblock In \emph{Workshop on Efficient Reasoning at the Conference on
  Language Modeling (COLM)}, 2026.

\bibitem[Gan et~al.(2026)Gan, Li, Wang, Meng, Wang, Zhao, Huo, and
  Gao]{gan2026raopd}
Siyuan Gan, Yuhan Li, Xiran Wang, Linjian Meng, Boyan Wang, Zhen Zhao, Jing
  Huo, and Yang Gao.
\newblock When teacher guidance misleads: Reward-aligned on-policy
  distillation.
\newblock \emph{arXiv preprint arXiv:2608.27960}, 2026.

\bibitem[Huang et~al.(2026)Huang, Liu, Goswami, Li, Jana, Kanakaris,
  Bl{\"o}baum, and Jain]{huang2026offpolicyteacher}
Langlin Huang, Hao Liu, Mononito Goswami, Xinyu Li, Prithwish Jana, Nikos
  Kanakaris, Patrick Bl{\"o}baum, and Purak Jain.
\newblock On the off-policy teacher in on-policy distillation.
\newblock \emph{arXiv preprint arXiv:2609.38360}, 2026.

\bibitem[Lai et~al.(2026)Lai, Li, Li, Liu, Li, and Zhao]{minio32025}
Xin Lai, Junyi Li, Wei Li, Tao Liu, Tianjian Li, and Hengshuang Zhao.
\newblock {Mini-o3}: Scaling up reasoning patterns and interaction turns for
  visual search.
\newblock In \emph{International Conference on Learning Representations
  (ICLR)}, 2026.

\bibitem[Shen et~al.(2025)Shen, Zhao, Zhao, Xu, Zhang, Zhu, and
  Yin]{zoomeye2024}
Haozhan Shen, Kangjia Zhao, Tiancheng Zhao, Ruochen Xu, Zilun Zhang, Mingwei
  Zhu, and Jianwei Yin.
\newblock {ZoomEye}: Enhancing multimodal {LLMs} with human-like zooming
  capabilities through tree-based image exploration.
\newblock In \emph{Proceedings of the 2025 Conference on Empirical Methods in
  Natural Language Processing}, pages 6602--6618, 2025.

\bibitem[Sheng et~al.(2025)Sheng, Zhang, Ye, Wu, Zhang, Zhang, Peng, Lin, and
  Wu]{verl2025}
Guangming Sheng, Chi Zhang, Zilingfeng Ye, Xibin Wu, Wang Zhang, Ru Zhang,
  Yanghua Peng, Haibin Lin, and Chuan Wu.
\newblock {HybridFlow}: A flexible and efficient {RLHF} framework.
\newblock In \emph{Proceedings of the Twentieth European Conference on Computer
  Systems}, pages 1279--1297, 2025.

\bibitem[Su et~al.(2025)Su, Wang, Ren, Lin, and Chen]{pixelreasoner2025}
Alex Su, Haozhe Wang, Weiming Ren, Fangzhen Lin, and Wenhu Chen.
\newblock {Pixel Reasoner}: Incentivizing pixel space reasoning via
  curiosity-driven reinforcement learning.
\newblock In \emph{Advances in Neural Information Processing Systems}, pages
  8222--8251, 2025.

\bibitem[Wang et~al.(2024)Wang, Bai, Tan, Wang, Fan, Bai, Chen, Liu, Wang, Ge,
  et~al.]{qwen2vl}
Peng Wang, Shuai Bai, Sinan Tan, Shijie Wang, Zhihao Fan, Jinze Bai, Keqin
  Chen, Xuejing Liu, Jialin Wang, Wenbin Ge, et~al.
\newblock {Qwen2-VL}: Enhancing vision-language model's perception of the world
  at any resolution.
\newblock \emph{arXiv preprint arXiv:2409.12191}, 2024.

\bibitem[Yang et~al.(2025)Yang, Kautz, and Hatamizadeh]{gateddeltanet2025}
Songlin Yang, Jan Kautz, and Ali Hatamizadeh.
\newblock Gated delta networks: Improving {Mamba2} with delta rule.
\newblock In \emph{International Conference on Learning Representations
  (ICLR)}, 2025.

\bibitem[Zhang et~al.(2025)Zhang, Khayatkhoei, Chhikara, and
  Ilievski]{vicrop2025}
Jiarui Zhang, Mahyar Khayatkhoei, Prateek Chhikara, and Filip Ilievski.
\newblock {MLLMs} know where to look: Training-free perception of small visual
  details with multimodal {LLMs}.
\newblock In \emph{International Conference on Learning Representations
  (ICLR)}, 2025.

\end{thebibliography}

%% file: sec/0_abstract.tex
\begin{abstract}
On-policy distillation (OPD) has emerged as an effective post-training approach to improving fine-grained perception in multimodal large language models by amortizing costly inference-time zooming operations into a single-pass recipe. However, vanilla OPD is inherently susceptible to
linguistic momentum: the teacher can follow the student's erroneous prefixes, thus weakening corrective supervision. To address this, we propose Blink On-Policy Distillation (BlinkOPD), in which the teacher ``blinks'' at every token: it reads the same prefix once with its eyes open and once with its eyes closed. A first-glance gain further strengthens the correction on rollouts that conflict with the teacher. By extracting more effective supervision from each on-policy trajectory, BlinkOPD achieves state-of-the-art average performance across six fine-grained benchmarks using only \textbf{500 training images (12× less data)} than existing OPD methods. Beyond fine-grained perception, BlinkOPD also demonstrates strong \textbf{zero-shot improvements} across \textbf{ten} held-out benchmarks spanning general visual perception, visual hallucination, and multimodal reasoning abilities. Code: \url{https://github.com/Supltz/BlinkOPD}%
\end{abstract}

%% file: sec/1_intro.tex
\section{Introduction}
\label{sec:intro}

\begin{figure}[t]
\centering
\input{figs/ability}
\caption{Comparison of existing OPSD methods on four visual abilities across 16 benchmarks using Qwen3.5-4B as the base model. With 12$\times$ fewer training images, \method outperforms competing methods on four visual abilities across 16 benchmarks.}
\label{fig:ability}
\end{figure}
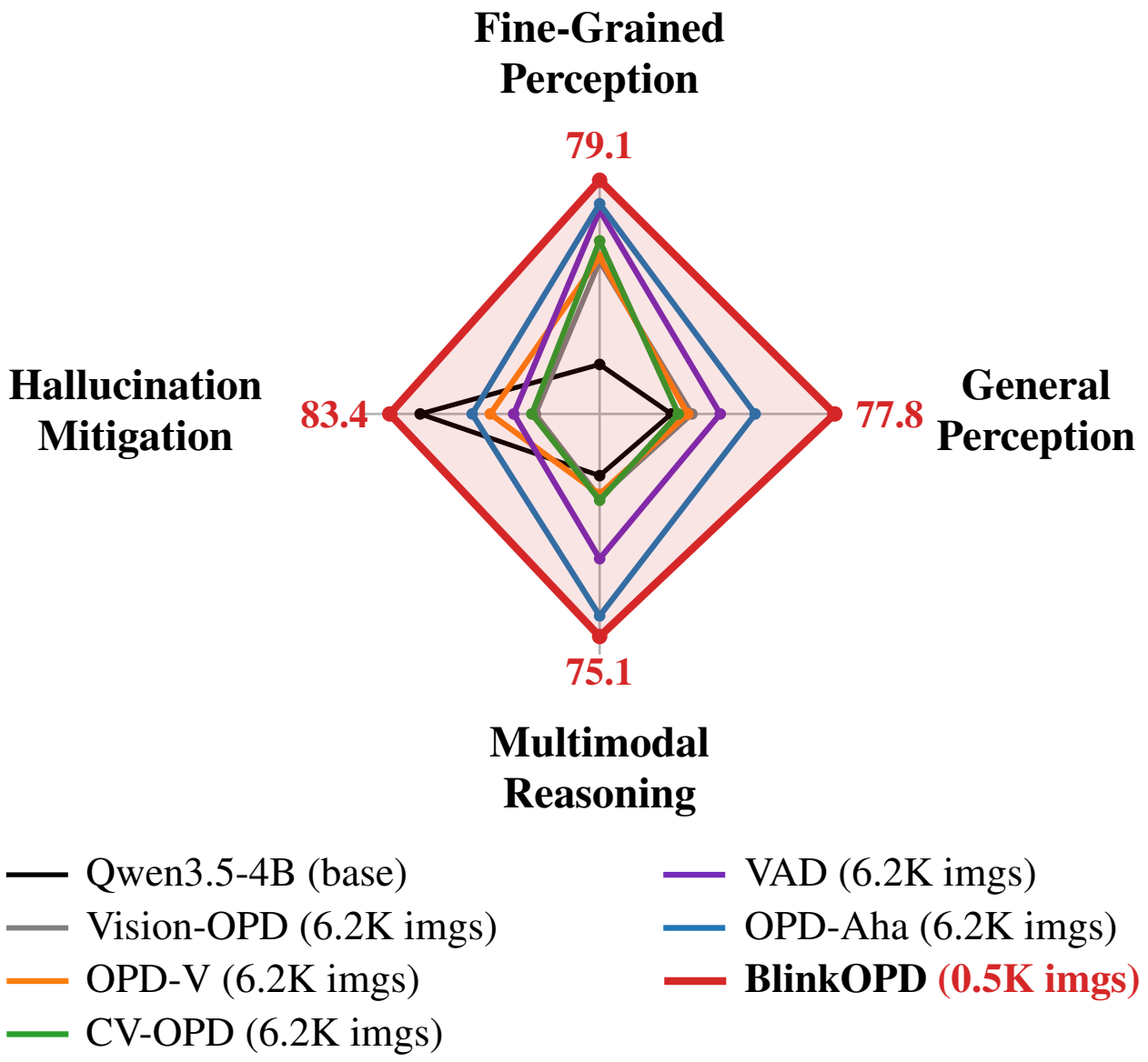

Multimodal large language models (MLLMs) excel at a wide range of visual understanding tasks, yet they still struggle with fine-grained perception when critical visual evidence is small or difficult to discern, such as distant signs or tiny text embedded in complex scenes~\cite{wu2024vstar,wang2025hrbench,zhang2025mmerealworld,zwz2026}. A prominent line of research, inspired by ``thinking with images''~\cite{o3thinking2025}, addresses this limitation by allocating additional inference-time computation to progressively inspect visual details through multi-turn cropping and zooming operations over regions of the original input image~\cite{deepeyes2025,thyme2025,deepeyesv2_2026,sensenovamars2026}.

An alternative line of research seeks to internalize fine-grained visual perception capabilities into the MLLM itself during post-training, so that subtle details can be recognized in a single forward pass \cite{chen2025shallower,barrios2026moda,zwz2026,visionopd2026,aha2026,vad2026,ead2026,cvopd2026,opdv2026,vzero2026,rpopsd2026,vicur2026}. Among them, \emph{on-policy distillation} (OPD) has recently emerged as a simple and effective post-training paradigm for internalizing visual zooming~\cite{visionopd2026,aha2026,vad2026,ead2026,vzero2026,rpopsd2026}. A straightforward approach is to distill from a teacher with stronger fine-grained visual perception capabilities~\cite{vzero2026,rpopsd2026}. However, such a teacher may not always be available, especially for frontier models. This has motivated growing interest in \emph{on-policy self-distillation} (OPSD), where a copy of the student itself, with access to zoomed crops of the target region as privileged visual inputs, serves as the teacher during training~\cite{visionopd2026,aha2026,vad2026,ead2026}.

A representative example is Vision-OPD~\cite{visionopd2026}, which distills privileged visual information from zoomed crops into a full-image student. Despite this visual privilege, its distillation target follows vanilla OPD~\cite{agarwal2024gkd}: the teacher's token-level predictions conditioned on student-generated prefixes serve directly as supervision targets. However, this vanilla conditioning objective is susceptible to \emph{linguistic momentum}: erroneous student prefixes can bias the teacher toward the student's own mistakes, thus weakening corrective supervision~\cite{aha2026}. This effect is particularly harmful to OPSD for fine-grained multimodal perception, as linguistic momentum from erroneous student prefixes can progressively override privileged visual evidence (see \cref{sec:capture}). 

To address this, contrastive methods reconstruct the distillation target using two predictions: a positive prediction from the privileged teacher conditioned on the zoomed crop, and a reference prediction obtained under altered or reduced visual input~\cite{aha2026,vcsd2026,vad2026,decomposedopd2026}. The key idea is to cancel the linguistic drift shared by both predictions, thereby isolating and amplifying privileged visual evidence to strengthen corrective supervision. Existing approaches construct such references using gray images~\cite{aha2026}, black images~\cite{vcsd2026,opdv2026}, Gaussian-noised images~\cite{leng2024vcd}, mismatched images~\cite{cvopd2026}, and text-only inputs~\cite{decomposedopd2026}. However, our analysis shows that image-based references, including gray, black, noisy, and mismatched images, fail to fully reproduce the prefix-induced linguistic drift, leaving substantial residual drift that limits their ability to counteract linguistic momentum. They also introduce a false belief that the queried object is absent. The text-only reference largely eliminates the residual drift, but the false belief still remains (see \cref{sec:refproblem}).

To overcome these limitations, we propose \emph{Blink On-Policy Distillation} (\method), which constructs an eyes-closed reference by letting the privileged teacher ``blink'' at every token. Specifically, the teacher processes each student prefix twice: once with its eyes open on the zoomed crop and once with its eyes closed, where the image-content tokens are removed while preserving the vision markers, chat template, and original positional indices. This design more effectively cancels prefix-induced linguistic drift and introduces less false beliefs. We further introduce a \emph{first-glance gain}: before training, the teacher takes a first glance at the zoomed crop without any prefix, and rollouts whose answers conflict with this first glance, which are likely to require stronger correction, receive a larger gain. 

To summarize, our contributions are as follows:

(1) We present a comprehensive analysis of existing OPSD distillation targets in fine-grained multimodal perception, showing that linguistic momentum overrides the privileged visual evidence, and that existing contrastive references leave residual drift and introduce false beliefs.

(2) We propose \method, which constructs an eyes-closed reference by removing image-content tokens while preserving the original multimodal input structure. Combined with a first-glance gain that selectively strengthens correction on potentially erroneous student trajectories, our method provides more effective corrective supervision.

(3) With only 500 training images, \method achieves strong performance across six fine-grained perception benchmarks, outperforming existing OPSD methods trained on about 12 times more images. It also demonstrates strong zero-shot transfer across ten held-out perception, hallucination, and multimodal reasoning benchmarks.

%% file: figs/ability.tex
\begin{tikzpicture}[font=\footnotesize]
\draw[black!30, line width=0.5pt] (0,0) -- (0.000,1.750);
\draw[black!30, line width=0.5pt] (0,0) -- (1.750,0.000);
\draw[black!30, line width=0.5pt] (0,0) -- (0.000,-1.750);
\draw[black!30, line width=0.5pt] (0,0) -- (-1.750,0.000);
\node[anchor=south, align=center, inner sep=1pt, font=\fontsize{9}{10.5}\selectfont\bfseries] (axFG) at (0.000,2.230) {Fine-Grained\\Perception};
\node[anchor=west, align=center, inner sep=1pt, font=\fontsize{9}{10.5}\selectfont\bfseries] (axGP) at (2.410,0.000) {General\\Perception};
\node[anchor=north, align=center, inner sep=1pt, font=\fontsize{9}{10.5}\selectfont\bfseries] (axRE) at (0.000,-2.230) {Multimodal\\Reasoning};
\node[anchor=east, align=center, inner sep=1pt, font=\fontsize{9}{10.5}\selectfont\bfseries] (axHA) at (-2.410,0.000) {Hallucination\\Mitigation};
\draw[draw=black, line width=1.0pt, line join=round] (0.000,0.360) -- (0.521,0.000) -- (0.000,-0.449) -- (-1.305,0.000) -- cycle;
\fill[black] (0.000,0.360) circle (1.2pt);
\fill[black] (0.521,0.000) circle (1.2pt);
\fill[black] (0.000,-0.449) circle (1.2pt);
\fill[black] (-1.305,0.000) circle (1.2pt);
\draw[draw=cVOPD, line width=1.1pt, line join=round] (0.000,1.110) -- (0.671,0.000) -- (0.000,-0.608) -- (-0.467,0.000) -- cycle;
\fill[cVOPD] (0.000,1.110) circle (1.2pt);
\fill[cVOPD] (0.671,0.000) circle (1.2pt);
\fill[cVOPD] (0.000,-0.608) circle (1.2pt);
\fill[cVOPD] (-0.467,0.000) circle (1.2pt);
\draw[draw=cOPDV, line width=1.2pt, line join=round] (0.000,1.140) -- (0.640,0.000) -- (0.000,-0.584) -- (-0.795,0.000) -- cycle;
\fill[cOPDV] (0.000,1.140) circle (1.2pt);
\fill[cOPDV] (0.640,0.000) circle (1.2pt);
\fill[cOPDV] (0.000,-0.584) circle (1.2pt);
\fill[cOPDV] (-0.795,0.000) circle (1.2pt);
\draw[draw=cCVOPD, line width=1.2pt, line join=round] (0.000,1.259) -- (0.572,0.000) -- (0.000,-0.628) -- (-0.493,0.000) -- cycle;
\fill[cCVOPD] (0.000,1.259) circle (1.2pt);
\fill[cCVOPD] (0.572,0.000) circle (1.2pt);
\fill[cCVOPD] (0.000,-0.628) circle (1.2pt);
\fill[cCVOPD] (-0.493,0.000) circle (1.2pt);
\draw[draw=cVAD, line width=1.2pt, line join=round] (0.000,1.481) -- (0.877,0.000) -- (0.000,-1.053) -- (-0.625,0.000) -- cycle;
\fill[cVAD] (0.000,1.481) circle (1.2pt);
\fill[cVAD] (0.877,0.000) circle (1.2pt);
\fill[cVAD] (0.000,-1.053) circle (1.2pt);
\fill[cVAD] (-0.625,0.000) circle (1.2pt);
\draw[draw=cAha, line width=1.2pt, line join=round] (0.000,1.529) -- (1.131,0.000) -- (0.000,-1.469) -- (-0.927,0.000) -- cycle;
\fill[cAha] (0.000,1.529) circle (1.2pt);
\fill[cAha] (1.131,0.000) circle (1.2pt);
\fill[cAha] (0.000,-1.469) circle (1.2pt);
\fill[cAha] (-0.927,0.000) circle (1.2pt);
\draw[draw=cOurs, line width=1.6pt, fill=cOurs, fill opacity=0.12, line join=round] (0.000,1.700) -- (1.710,0.000) -- (0.000,-1.619) -- (-1.526,0.000) -- cycle;
\fill[cOurs] (0.000,1.700) circle (1.6pt);
\fill[cOurs] (1.710,0.000) circle (1.6pt);
\fill[cOurs] (0.000,-1.619) circle (1.6pt);
\fill[cOurs] (-1.526,0.000) circle (1.6pt);
\node[anchor=south, font=\footnotesize\bfseries, text=cOurs, inner sep=0.8pt] at (0.000,1.820) {79.1};
\node[anchor=west, font=\footnotesize\bfseries, text=cOurs, inner sep=0.8pt] at (1.830,0.000) {77.8};
\node[anchor=north, font=\footnotesize\bfseries, text=cOurs, inner sep=0.8pt] at (0.000,-1.739) {75.1};
\node[anchor=east, font=\footnotesize\bfseries, text=cOurs, inner sep=0.8pt] at (-1.646,0.000) {83.4};
\node[anchor=north west, inner sep=0pt, font=\fontsize{8.5}{10}\selectfont] at (axHA.west |- 0,-3.180) {\begin{tabular}{@{}l@{\hspace{3pt}}l@{}}\tikz[baseline=-0.6ex]\draw[draw=black, line width=1.0pt] (0,0) -- (0.45,0); & Qwen3.5-4B (base) \\[1pt] \tikz[baseline=-0.6ex]\draw[draw=cVOPD, line width=1.1pt] (0,0) -- (0.45,0); & Vision-OPD (6.2K imgs) \\[1pt] \tikz[baseline=-0.6ex]\draw[draw=cOPDV, line width=1.2pt] (0,0) -- (0.45,0); & OPD-V (6.2K imgs) \\[1pt] \tikz[baseline=-0.6ex]\draw[draw=cCVOPD, line width=1.2pt] (0,0) -- (0.45,0); & CV-OPD (6.2K imgs)\end{tabular}};
\node[anchor=north east, inner sep=0pt, font=\fontsize{8.5}{10}\selectfont] at (axGP.east |- 0,-3.180) {\begin{tabular}{@{}l@{\hspace{3pt}}l@{}}\tikz[baseline=-0.6ex]\draw[draw=cVAD, line width=1.2pt] (0,0) -- (0.45,0); & VAD (6.2K imgs) \\[1pt] \tikz[baseline=-0.6ex]\draw[draw=cAha, line width=1.2pt] (0,0) -- (0.45,0); & OPD-Aha (6.2K imgs) \\[1pt] \tikz[baseline=-0.6ex]\draw[draw=cOurs, line width=1.6pt] (0,0) -- (0.45,0); & \textbf{\methodfig} \textcolor{cOurs}{\textbf{(0.5K imgs)}}\end{tabular}};
\end{tikzpicture}

%% file: sec/3_analysis.tex
\section{Preliminaries and Analysis}
\label{sec:analysis}

\subsection{On-Policy Self-Distillation}
\label{sec:prelim}
In vanilla OPSD for multimodal fine-grained perception \cite{visionopd2026}, the student $\pi_\theta$ receives the full image $I$ and question $q$, denoted by $x_s=(I,q)$, while the teacher is initialized as a copy of the student and receives privileged visual information during training. In our setting, the privileged input is a zoomed crop $I^+$ of the target region, giving $x_T=(I^+,q)$.  The student generates on-policy rollouts $y\sim\pi_\theta(\cdot\mid x_s)$, and the teacher scores each student-generated prefix $y_{<t}$ to provide token-level supervision without requiring ground-truth answers. For each prefix, the teacher predicts a next-token distribution over the vocabulary $\pT_t=p_\phi(\cdot\mid x_T,y_{<t})$. The student is then trained to match a distillation target distribution $q_t$, which in vanilla OPSD is the teacher prediction itself, $q_t=\pT_t$, by minimizing
\begin{equation}
\mathcal{L}_{\text{OPD}}
=\mathbb{E}_{\scalebox{1.25}{$\scriptstyle y\sim\pi_\theta(\cdot\mid x_s)$}}
\Bigg[
\frac{1}{|y|}
\sum_{t=1}^{|y|}
\mathcal{D}\big(
\pi_\theta(\cdot\mid x_s,y_{<t})
\,\big\|\,q_t
\big)
\Bigg],
\label{eq:opd}
\end{equation}
where $\mathcal{D}(\cdot\,\|\,\cdot)$ denotes a distribution-matching divergence.

\subsection{Linguistic Momentum in OPSD}
\label{sec:capture}
\begin{figure}[t]
\centering
\input{figs/capture}
\caption[Linguistic momentum probe]{Comparison of the distillation targets of Vision-OPD, VAD, CV-OPD, OPD-V, OPD-Aha and our proposed \method on linguistic momentum and privileged evidence survival.
(a) As the teacher reads more of the wrong prefix\protect\footnotemark, the follow rate tends to rise: Vision-OPD rises from 23\% to 100\%, VAD, CV-OPD and OPD-V from about 50\% to 100\%, and OPD-Aha still ends at 64\%, whereas \method stays below 40\% throughout.
(b) As the teacher reads more of the wrong prefix, the survival score tends to decrease: Vision-OPD, VAD, CV-OPD and OPD-V turn negative within the first quarter of the prefix and never recover, and OPD-Aha also falls below zero by the end, whereas \method remains positive and outperforms the others throughout.}
\label{fig:capture}
\end{figure}
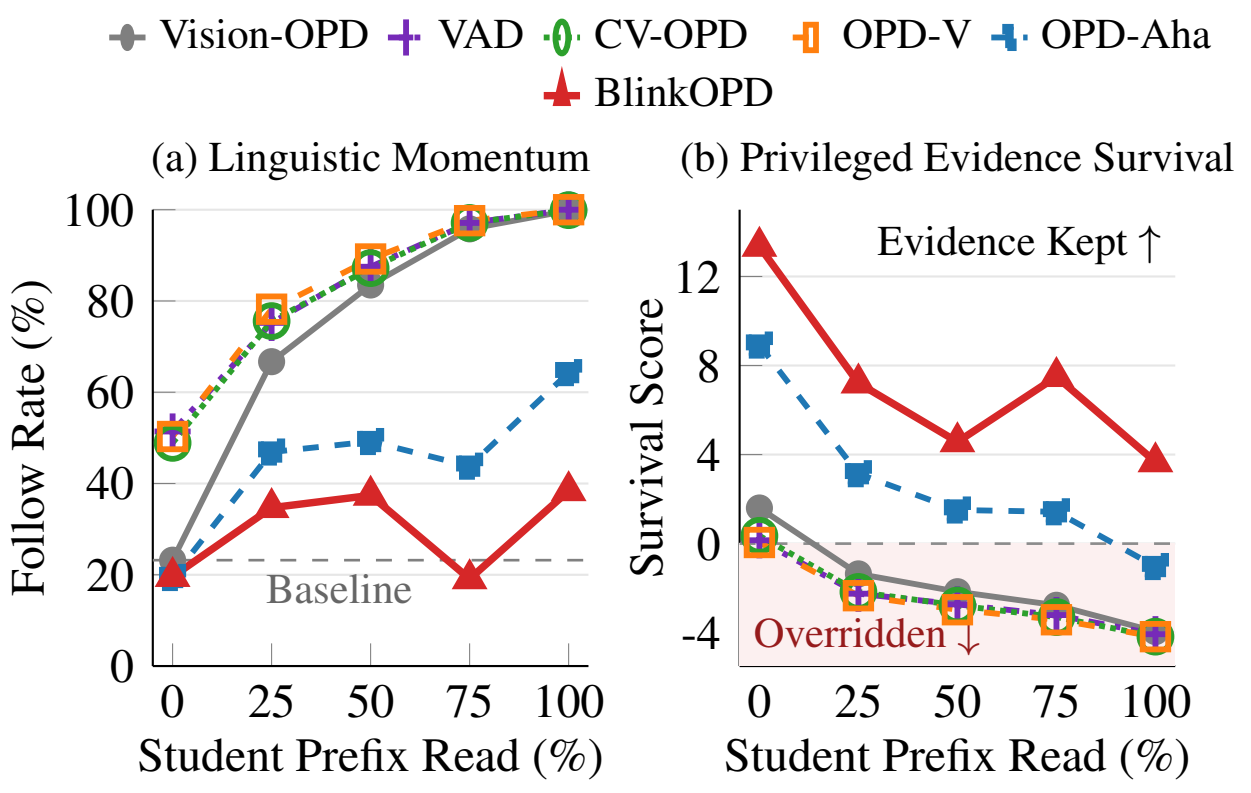

We design a controlled experiment to investigate linguistic momentum and privileged evidence survival in OPSD for fine-grained multimodal perception. Specifically, we randomly sample 768 questions from the Vision-OPD \cite{visionopd2026} training pool and use Qwen3.5-4B to generate four rollouts per question from the full-image input. We truncate each rollout at 0, 25, 50, 75, and 100\% of its tokens to observe the teacher at different stages of the student's rollout. For each truncated prefix, the privileged teacher observes the zoomed crop, and we record its next-token probabilities over the answer options. Using these read-outs, we construct the distillation targets of Vision-OPD \cite{visionopd2026}, VAD \cite{vad2026}, CV-OPD \cite{cvopd2026}, OPD-V \cite{opdv2026}, OPD-Aha \cite{aha2026}, and our proposed \method to compare their susceptibility to linguistic momentum and privileged evidence survival, as shown in \cref{fig:capture}. Full experimental details are provided in \textbf{Appendix \textcolor{red}{A}}.

\noindent\textbf{Follow Rate.}
To measure how often the supervision endorses the student's own mistake, \ie, the strength of linguistic momentum, we define the follow rate.
Let $\mathcal{W}$ be the set of student rollouts whose final answer is wrong; for $y\in\mathcal{W}$, let $a_y$ be the option chosen by the student.
For a method and a truncation point $\tau$, let $q^{\tau}_y$ denote its distillation target $q_t$ after the first $\tau$ of $y$, read out over the set $\mathcal{O}_y$ of answer options of the question. The follow rate is
\begin{equation}
F(\tau)=\frac{1}{|\mathcal{W}|}\sum_{y\in\mathcal{W}}\mathbf{1}\Big[\argmax_{\mbox{\small$o\,{\in}\,\mathcal{O}_y$}}\,q^{\tau}_y(o)=a_y\Big],
\label{eq:follow}
\end{equation}
\footnotetext{We also provide a control experiment that replaces the prefix with another question's answer of the same length, which confirms that this rise comes from the content of the prefix rather than its length (\textbf{Appendix \textcolor{red}{B}}).}
\ie, the fraction of wrong rollouts on which the distillation target picks the student's own wrong option.
Training the student toward a distillation target with a high follow rate reinforces the student's error.
At $\tau=0$, the teacher has not yet read the student's answer, so $F(0)$ serves as a baseline: it only counts cases in which the teacher happens to make the same mistake as the student. Any rise of $F$ above $F(0)$ is caused by reading the student's prefix, and this rise is what measures linguistic momentum.

\noindent\textbf{Survival Score.}
To measure how much of the privileged visual evidence survives in the distillation target after the teacher reads the student's wrong prefix, we define the survival score.
Let $g_y\in\mathcal{O}_y$ be the correct option of rollout $y$, and let $\mathcal{K}\subseteq\mathcal{W}$ be the wrong rollouts on which the teacher picks $g_y$ before reading the prefix.
On these rollouts, the crop holds the evidence that the student missed, and at $\tau=0$ the teacher's preference for $g_y$ over $a_y$ reflects this privileged visual evidence.
The survival score is
\begin{equation}
S(\tau)=\frac{1}{|\mathcal{K}|}\sum_{y\in\mathcal{K}}\log\frac{q^{\tau}_y(g_y)}{q^{\tau}_y(a_y)}.
\label{eq:margin}
\end{equation}
A positive value means that the distillation target still conveys the privileged evidence, and a negative value means that the evidence is overridden by the student's prefix.

\subsection{Limitations of Existing Contrastive References}
\label{sec:refproblem}
\begin{figure}[t]
\centering
\input{figs/refmap}
\caption{Comparison of contrastive references on residual drift and false belief.
(a) Image-based references leave 20--23\% of the drift uncancelled, whereas the text-only and eyes-closed references leave about 10\%.
(b) All other references raise the probability of ``not present'' false belief to 0.39--0.63. Only our proposed eyes-closed reference is low on both measures.}
\label{fig:refmap}
\end{figure}
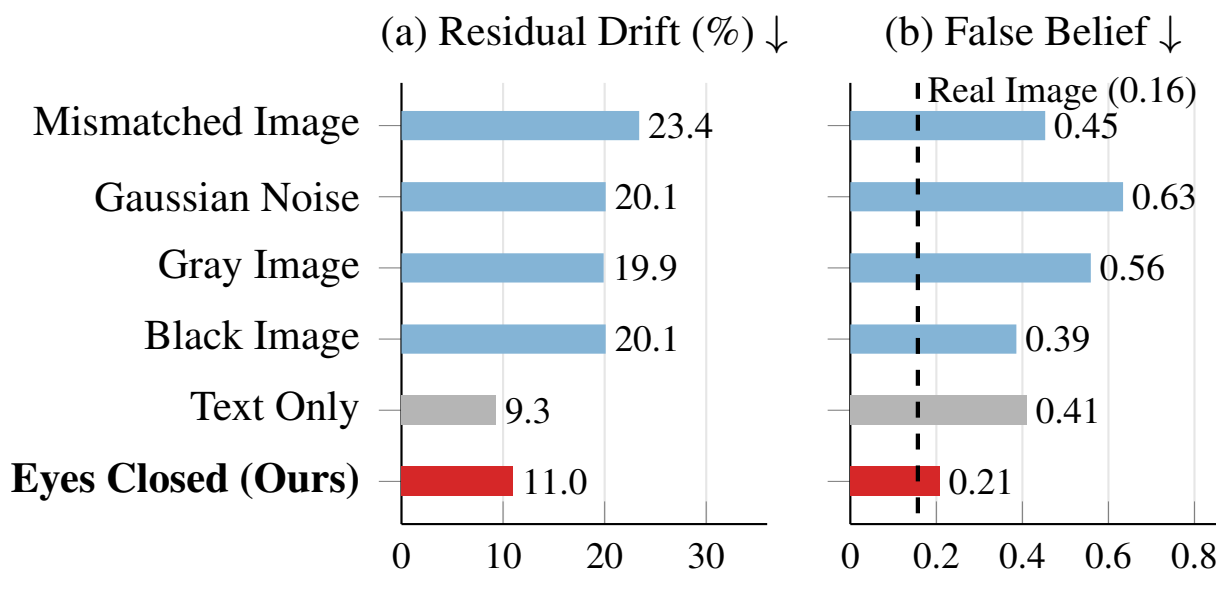
Existing contrastive methods~\cite{aha2026,vad2026,vcsd2026,opdv2026,cvopd2026,decomposedopd2026,leng2024vcd} counter linguistic momentum by comparing the privileged teacher with a reference that reads the same student prefix.
Given a reference distribution $\pR_t$, the distillation target sharpens the teacher as
\begin{equation}
q^{\beta}_t=\operatorname{softmax}\!\Big(\log\pT_t+\beta\,\underbrace{\big[\log\pT_t-\log\pR_t\big]}_{\mbox{\small$r_t$}}\Big),
\label{eq:tilt}
\end{equation}
where $r_t$ compares the teacher with the reference token by token and the gain $\beta\ge0$ sets how strongly the distillation target is tilted toward tokens with a high $r_t$.

Ideally, since both distributions read the same student prefix, the part of the prediction induced by the prefix alone is shared and cancels in $r_t$, while the part that comes from the privileged crop remains.
Whether this happens, however, depends entirely on the reference, which must meet two requirements.
First, it must be drifted by the student prefix exactly as much as the teacher, so that no \emph{residual drift} remains in $r_t$.
Second, while holding no privileged visual evidence, it must not add a false belief that the queried object is absent which enters the distillation target through $r_t$.
Existing methods ~\cite{aha2026,vcsd2026,opdv2026,leng2024vcd,cvopd2026,decomposedopd2026} build the reference either by replacing the crop with another image (gray, black, noise or a mismatched image) or by removing the image from the conversation.
Using the probe of \cref{sec:capture}, we test whether these references meet these requirements.

\noindent\textbf{Residual Drift.}
We define the residual drift of a reference to measure how much of the prefix-induced drift it fails to cancel.
Let $S_{+}(\tau)$ and $S_{\mathrm{ref}}(\tau)$ be the survival scores of the teacher and of the reference, \ie, \cref{eq:margin} with $q^{\tau}_y$ replaced by their predictions.
How far the teacher is drifted by the whole student prefix is then the fall of its survival score, $\Delta_{+}=S_{+}^{\,0}-S_{+}^{\,\mathrm{end}},$ where the superscripts $0$ and $\mathrm{end}$ denote the values before reading the prefix and after reading the whole student prefix, \ie, $\tau=0$ and $100\%$\footnote{We also report $\delta$ at $\tau = 50\% \;\text{and}\; 75\% $. At both end points, image-based references leave more drift than the text-only and eyes-closed references, giving the same picture as the whole prefix (\textbf{Appendix \textcolor{red}{C}}).}.
The reference reads the same prefix and is dragged by $\Delta_{\mathrm{ref}}=S_{\mathrm{ref}}^{\,0}-S_{\mathrm{ref}}^{\,\mathrm{end}}$.
The residual drift is the share of the teacher's drift that the reference still leaves uncancelled in the contrast $r_t$:
\begin{equation}
\delta=\frac{\Delta_{+}-\Delta_{\mathrm{ref}}}{\Delta_{+}} .
\label{eq:kappa}
\end{equation}
$\delta$ is lower the better with $\delta=0$ meaning that the reference is drifted exactly as much as the teacher and are all cancelled.
\noindent\textbf{Proposition 1.} $\delta$ in \cref{eq:kappa} is exactly the residual drift of the contrast $r_t$ in \cref{eq:tilt} (see proof in \textbf{Appendix \textcolor{red}{D}}).

\noindent\textbf{False Belief.}
In \cref{fig:refmap}a, the text-only reference leaves the least residual drift.
However, it still introduces a false ``not present'' belief that the teacher does not have.
We take 300 multiple-choice questions from MME-RealWorld-CN~\cite{zhang2025mmerealworld} whose options include ``not present / cannot be determined''.
For each reference, the teacher reads only the question, without any student prefix, and we record the probability that it assigns to this option at the first answer token (\cref{fig:refmap}b). The same procedure with the original real image of each question gives 0.16. A reference with a higher probability than 0.16 therefore introduces a false belief that the queried object is absent.
The false belief is itself a form of hallucination: a reference with a high false belief can introduce hallucination into the student (see the experimental results on hallucination benchmarks in \cref{sec:zeroshot}).

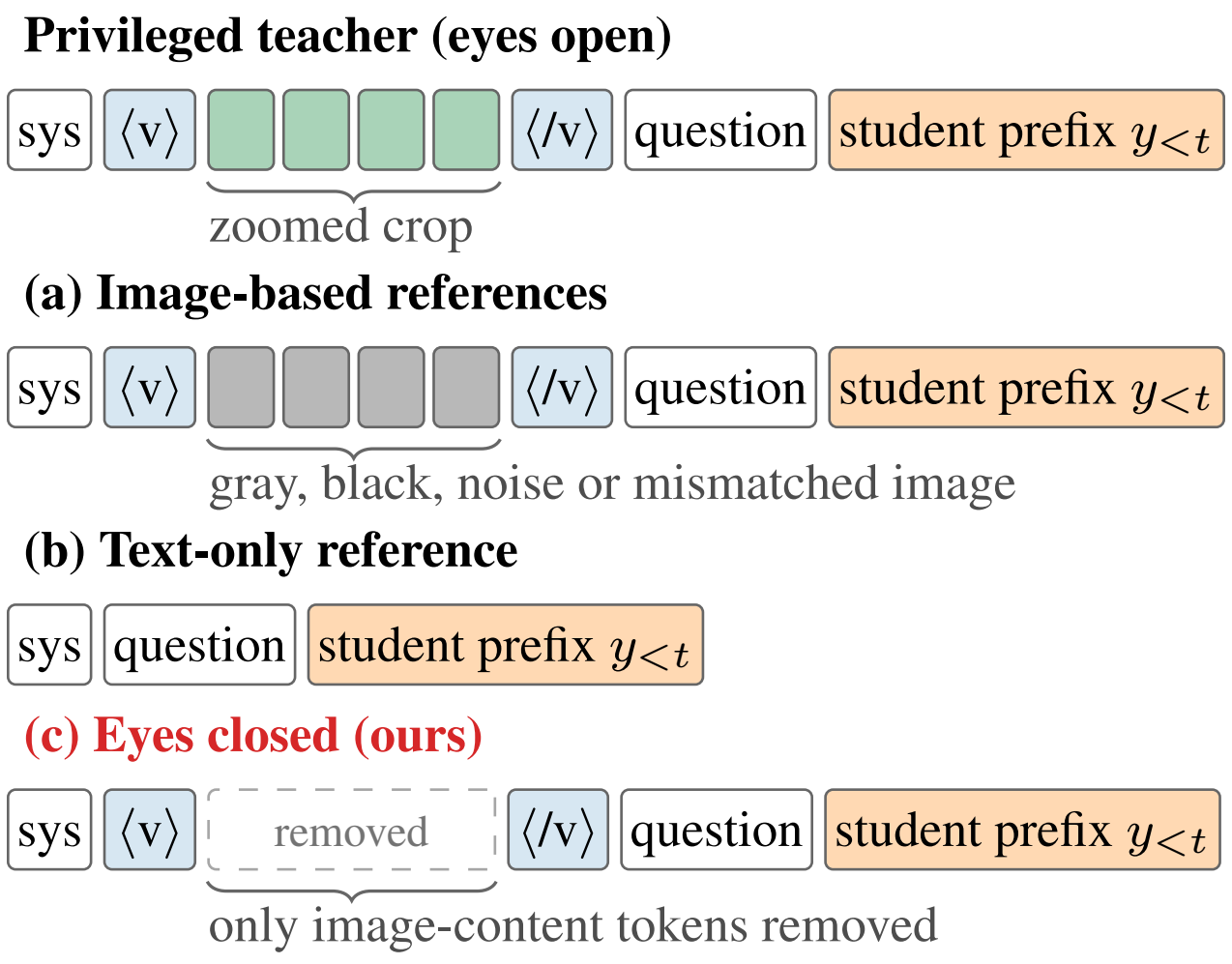
\begin{figure}[t]
\centering
\input{figs/teaser}
\caption{Contrastive reference construction for OPSD. (a) Image-based references replace the crop with gray, black, noisy, or mismatched images~\cite{aha2026,vcsd2026,opdv2026,leng2024vcd,cvopd2026}. (b) The text-only reference removes the image from the conversation~\cite{decomposedopd2026}. (c) Our eyes-closed reference removes only the image-content tokens while preserving vision markers and original positional indices.}
\label{fig:teaser}
\end{figure}

%% file: figs/capture.tex
\begin{tikzpicture}
\begin{groupplot}[
  group style={group size=2 by 1, horizontal sep=1.0cm},
  width=0.5355\linewidth, height=4.6cm,
  xmin=-5, xmax=105, xtick={0,25,50,75,100},
  xticklabels={0,25,50,75,100},
  xlabel={Student Prefix Read (\%)},
  xlabel style={font=\small, yshift=3pt},
  ticklabel style={font=\small},
  ylabel style={font=\small, yshift=-5pt},
  title style={font=\footnotesize, yshift=-4pt},
  axis lines*=left, ymajorgrids, grid style={black!10},
  legend style={font=\footnotesize, draw=none, fill=none, inner sep=0.5pt, /tikz/every even column/.append style={column sep=1pt}},
  legend image post style={xscale=0.4},
  legend cell align=left,
  every axis plot/.append style={line width=1.1pt, mark size=2.0pt},
]
\nextgroupplot[
  title={(a) Linguistic Momentum},
  ylabel={Follow Rate (\%)}, ymin=0, ymax=100, ytick={0,20,40,60,80,100},
  legend to name=capleg, legend columns=5,
]
\draw[black!50, dashed, line width=0.5pt] (axis cs:-5,23.2) -- (axis cs:105,23.2);
\node[font=\footnotesize, text=black!60, anchor=north west, inner sep=1pt] at (axis cs:22,21.5) {Baseline};
\addplot[color=cVOPD, mark=*] coordinates {(0,23.2) (25,66.7) (50,83.5) (75,95.8) (100,99.8)};
\addlegendentry{Vision-OPD}
\addplot[color=cVAD, mark=+, mark size=3.4pt, dashdotted, mark options={solid, line width=1.1pt}] coordinates {(0,51.4) (25,75.2) (50,87.5) (75,97.1) (100,100.0)};
\addlegendentry{VAD}
\addplot[color=cCVOPD, mark=o, mark size=3.0pt, densely dotted, mark options={solid, fill=white}] coordinates {(0,48.9) (25,75.6) (50,87.2) (75,97.1) (100,99.9)};
\addlegendentry{CV-OPD}
\addplot[color=cOPDV, mark=square, mark size=2.4pt, loosely dashed, mark options={solid, fill=white}] coordinates {(0,50.3) (25,78.2) (50,89.2) (75,97.6) (100,100.0)};
\addlegendentry{OPD-V}
\addplot[color=cAha, mark=square*, dashed] coordinates {(0,19.4) (25,46.9) (50,49.2) (75,43.8) (100,64.3)};
\addlegendentry{OPD-Aha}
\addlegendimage{empty legend}\addlegendentry{}
\addlegendimage{empty legend}\addlegendentry{}
\addplot[color=cOurs, mark=triangle*, line width=1.4pt, mark size=2.5pt] coordinates {(0,19.6) (25,34.7) (50,37.4) (75,19.1) (100,38.4)};
\addlegendentry{\methodfig}
\nextgroupplot[
  title={(b) Privileged Evidence Survival},
  ylabel={Survival Score}, ymin=-5.5, ymax=15, ytick={-4,0,4,8,12},
]
\fill[cOurs!7] (axis cs:-5,-5.5) rectangle (axis cs:105,0);
\draw[black!45, dashed, line width=0.5pt] (axis cs:-5,0) -- (axis cs:105,0);
\addplot[color=cVOPD, mark=*] coordinates {(0,1.60) (25,-1.36) (50,-2.16) (75,-2.77) (100,-3.95)};
\addplot[color=cVAD, mark=+, mark size=3.4pt, dashdotted, mark options={solid, line width=1.1pt}] coordinates {(0,0.15) (25,-2.25) (50,-2.73) (75,-3.21) (100,-4.07)};
\addplot[color=cCVOPD, mark=o, mark size=3.0pt, densely dotted, mark options={solid, fill=white}] coordinates {(0,0.35) (25,-2.17) (50,-2.78) (75,-3.30) (100,-4.18)};
\addplot[color=cOPDV, mark=square, mark size=2.4pt, loosely dashed, mark options={solid, fill=white}] coordinates {(0,0.05) (25,-2.34) (50,-2.97) (75,-3.42) (100,-4.18)};
\addplot[color=cAha, mark=square*, dashed] coordinates {(0,8.92) (25,3.14) (50,1.51) (75,1.43) (100,-1.03)};
\addplot[color=cOurs, mark=triangle*, line width=1.4pt, mark size=2.5pt] coordinates {(0,13.35) (25,7.21) (50,4.57) (75,7.50) (100,3.67)};
\node[font=\footnotesize, text=black, anchor=north east, align=right, inner sep=1pt] at (axis cs:103,14.6) {Evidence Kept $\uparrow$};
\node[font=\footnotesize, text=cOurs!70!black, anchor=south west, align=left, inner sep=1pt] at (axis cs:-3,-5.3) {Overridden $\downarrow$};
\end{groupplot}
\node[anchor=south] at ($(group c1r1.north)!0.5!(group c2r1.north)+(0,0.42cm)$) {\pgfplotslegendfromname{capleg}};
\end{tikzpicture}

%% file: figs/refmap.tex
\begin{tikzpicture}
\begin{groupplot}[
  group style={group size=2 by 1, horizontal sep=0.55cm, yticklabels at=edge left},
  width=0.48\linewidth, height=4.5cm,
  xbar, /pgf/bar width=5.5pt, /pgf/bar shift=0pt,
  y dir=reverse,
  ytick={1,2,3,4,5,6},
  yticklabels={Mismatched Image, Gaussian Noise, Gray Image, Black Image, Text Only, \textbf{Eyes Closed (Ours)}},
  yticklabel style={font=\footnotesize},
  ymin=0.4, ymax=6.6,
  xticklabel style={font=\scriptsize},
  title style={font=\footnotesize, yshift=-4pt},
  axis lines*=left, xmajorgrids, grid style={black!10},
  nodes near coords, nodes near coords style={font=\scriptsize, anchor=west, inner sep=1.5pt},
  every node near coord/.append style={/pgf/number format/.cd, fixed, fixed zerofill, precision=1},
  clip=false,
]
\nextgroupplot[title={(a) Residual Drift (\%) $\downarrow$}, xmin=0, xmax=36]
\addplot[fill=cAha!55, draw=none] coordinates {(23.4,1) (20.1,2) (19.9,3) (20.1,4)};
\addplot[fill=cGray!45, draw=none] coordinates {(9.3,5)};
\addplot[fill=cOurs, draw=none] coordinates {(11.0,6)};
\nextgroupplot[title={(b) False Belief $\downarrow$}, xmin=0, xmax=0.85,
  every node near coord/.append style={/pgf/number format/.cd, fixed, precision=2}]
\addplot[fill=cAha!55, draw=none] coordinates {(0.453,1) (0.634,2) (0.559,3) (0.386,4)};
\addplot[fill=cGray!45, draw=none] coordinates {(0.410,5)};
\addplot[fill=cOurs, draw=none] coordinates {(0.208,6)};
\draw[dashed, black, line width=0.8pt] (axis cs:0.157,0.4) -- (axis cs:0.157,6.6);
\node[font=\scriptsize, text=black, anchor=west, inner sep=1pt] at (axis cs:0.165,0.55) {Real Image (0.16)};
\end{groupplot}
\end{tikzpicture}

%% file: figs/teaser.tex
\resizebox{\linewidth}{!}{%
\begin{tikzpicture}[
  font=\footnotesize,
  tok/.style={draw=black!60, rounded corners=1.2pt, minimum height=4.4mm, inner sep=1.4pt, line width=0.4pt, text height=1.6ex, text depth=0.45ex},
  txt/.style={tok, fill=white},
  mrk/.style={tok, fill=cAha!18, minimum width=5.0mm},
  img/.style={tok, fill=ForestGreen!35, minimum width=3.6mm},
  gry/.style={tok, fill=black!28, minimum width=3.6mm},
  gone/.style={tok, draw=black!35, dashed, fill=white, minimum width=15.8mm},
  pre/.style={tok, fill=orange!30},
  lab/.style={font=\footnotesize\bfseries, anchor=west, inner sep=0pt},
  brc/.style={decorate, decoration={brace, mirror, amplitude=3pt}, line width=0.5pt, black!60},
  bl/.style={font=\footnotesize, text=black!70, anchor=north west, inner sep=0pt},
  >=Stealth]
\def\rowgap{1.42}
\node[lab] at (-0.15, 0.50) {Privileged teacher (eyes open)};
\node[txt] (s0) at (0,0) {sys};
\node[mrk, right=0.6mm of s0] (v0) {$\langle$v$\rangle$};
\node[img, right=0.6mm of v0] (i01) {};
\node[img, right=0.4mm of i01] (i02) {};
\node[img, right=0.4mm of i02] (i03) {};
\node[img, right=0.4mm of i03] (i04) {};
\node[mrk, right=0.6mm of i04] (w0) {$\langle$/v$\rangle$};
\node[txt, right=0.6mm of w0] (q0) {question};
\node[pre, right=0.6mm of q0] (y0) {student prefix $y_{<t}$};
\draw[brc] ([yshift=-0.6mm]i01.south west) -- ([yshift=-0.6mm]i04.south east);
\node[bl] at ([yshift=-1.9mm]i01.south west) {zoomed crop};
\begin{scope}[yshift=-\rowgap cm]
\node[lab] at (-0.15, 0.50) {(a) Image-based references};
\node[txt] (s1) at (0,0) {sys};
\node[mrk, right=0.6mm of s1] (v1) {$\langle$v$\rangle$};
\node[gry, right=0.6mm of v1] (i11) {};
\node[gry, right=0.4mm of i11] (i12) {};
\node[gry, right=0.4mm of i12] (i13) {};
\node[gry, right=0.4mm of i13] (i14) {};
\node[mrk, right=0.6mm of i14] (w1) {$\langle$/v$\rangle$};
\node[txt, right=0.6mm of w1] (q1) {question};
\node[pre, right=0.6mm of q1] (y1) {student prefix $y_{<t}$};
\draw[brc] ([yshift=-0.6mm]i11.south west) -- ([yshift=-0.6mm]i14.south east);
\node[bl] at ([yshift=-1.9mm]i11.south west) {gray, black, noise or mismatched image};
\end{scope}
\begin{scope}[yshift=-2*\rowgap cm]
\node[lab] at (-0.15, 0.50) {(b) Text-only reference};
\node[txt] (s2) at (0,0) {sys};
\node[txt, right=0.6mm of s2] (q2) {question};
\node[pre, right=0.6mm of q2] (y2) {student prefix $y_{<t}$};
\end{scope}
\begin{scope}[yshift=-2*\rowgap cm - 1.02cm]
\node[lab, text=cOurs] at (-0.15, 0.50) {(c) Eyes closed (ours)};
\node[txt] (s3) at (0,0) {sys};
\node[mrk, right=0.6mm of s3] (v3) {$\langle$v$\rangle$};
\node[gone, right=0.6mm of v3, font=\scriptsize, text=black!55] (i31) {removed};
\node[mrk, right=0.6mm of i31] (w3) {$\langle$/v$\rangle$};
\node[txt, right=0.6mm of w3] (q3) {question};
\node[pre, right=0.6mm of q3] (y3) {student prefix $y_{<t}$};
\draw[brc] ([yshift=-0.6mm]i31.south west) -- ([yshift=-0.6mm]i31.south east);
\node[bl] at ([yshift=-1.9mm]i31.south west) {only image-content tokens removed};
\end{scope}
\end{tikzpicture}}

%% file: sec/4_method.tex
\section{\methodlong}
\label{sec:method}

\subsection{Closing the Teacher's Eyes}
\label{sec:eyes}
As illustrated in \cref{fig:teaser}c, the eyes-closed reference is the privileged teacher with only image-content tokens removed: the vision markers, the chat template, the question and the prefix are kept, and every remaining token keeps the position it had in the full sequence,
\begin{equation}
\pC_t=p_{\phi}\big(\cdot\mid\texttt{sys},\langle\text{v}\rangle,\langle/\text{v}\rangle,q,y_{<t};\,\mathrm{pos}\big),
\label{eq:closed}
\end{equation}
where $\mathrm{pos}$ denotes these original positions; see \textbf{Appendix \textcolor{red}{E}} for the implementation details of this input in training.

\subsection{First-Glance Gain}
\label{sec:glance}
Even with a faithful reference, the contrastive target still depends on the gain $\beta$ in \cref{eq:tilt}.
The reference decides \emph{what} is amplified while the gain decides \emph{how much}.
We find that a single global gain does not suit all rollouts.
On a student-wrong rollout, the teacher is dragged by the prefix, and a large gain recovers the correct option.
On a student-correct rollout, however, a large gain pushes the distillation target away from the correct answer.
Our probe confirms this trade-off (see \textbf{Appendix \textcolor{red}{F}}), so the gain should be raised only on rollouts that are likely to be wrong.

Detecting such rollouts would normally require the correct answer.
The frozen teacher, however, offers a label-free signal.
Before training, the teacher looks at the crop once ($x_T$, no prefix) and records its answer $\hat a^{\text{glance}}(x)$, its \emph{first glance}.
A rollout $y_i$ conflicts with the glance when its final answer $\hat a(y_i)$ differs from it, and we raise its gain as:
\begin{equation}
\beta_i=\beta\,\gamma^{\,\scalebox{0.92}{$\mathbf{1}\big[\hat a(y_i)\neq\hat a^{\text{glance}}(x)\big]$}},
\label{eq:beta}
\end{equation}
where $\beta$ is the base gain of \cref{eq:tilt} and $\gamma>1$ scales it up on conflicting rollouts. See \textbf{Appendix \textcolor{red}{G}} for an experimental comparison of the global gain and the first-glance gain.

\subsection{Distillation Target and Training}
\label{sec:objective}
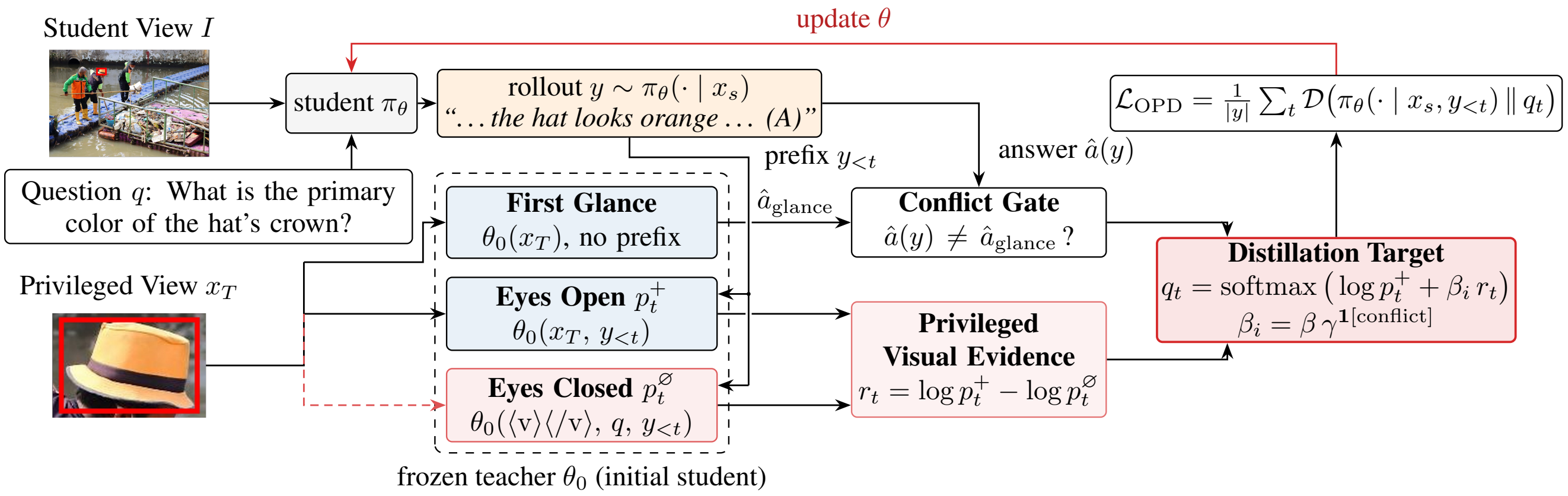
\begin{figure*}[t]
\centering
\resizebox{\textwidth}{!}{\input{figs/overview}}
\caption{Overview of \method. The student generates rollouts from the full image $I$ and question $q$, together denoted by $x_s$. The frozen privileged teacher $\theta_0$ reads each student prefix twice: once with its eyes open on the zoomed crop ($\pT_t$), and once with its eyes closed ($\pC_t$, \cref{sec:eyes}). Their log-ratio $r_t$ keeps the privileged visual evidence, while the prefix-induced linguistic drift shared by both cancels. A prefix-free first glance of the same teacher flags rollouts whose answer conflicts with it, and the gain is raised by a factor $\gamma$ on these rollouts (\cref{sec:glance}). The student is trained toward the resulting distillation target (\cref{sec:objective}). At test time, the student runs alone in a single pass.}
\label{fig:overview}
\end{figure*}

\method keeps the OPSD objective of \cref{eq:opd} and replaces only its distillation target $q_t$.
For rollout $y_i$, it combines the eyes-closed reference of \cref{eq:closed} with the gain of \cref{eq:beta}:
\begin{equation}
q_{i,t}=\operatorname{softmax}\Big(\log\pT_t+\beta_i\,\big[\log\pT_t-\log\pC_t\big]\Big).
\label{eq:target}
\end{equation}
\cref{fig:overview} provides an overview of \method. An algorithm chart for one training step is provided in \textbf{Appendix \textcolor{red}{H}}.
The entire \method training pipeline requires neither assistance from larger models nor ground-truth answers.

%% file: figs/overview.tex
\begin{tikzpicture}[
  >=Stealth, font=\normalsize,
  box/.style={draw=black, rounded corners=2pt, line width=0.5pt, align=center, inner sep=1.5pt},
  ph/.style={box, fill=black!6, draw=black, minimum width=2.0cm, minimum height=1.3cm},
  stu/.style={box, fill=black!4, minimum width=1.65cm, minimum height=0.75cm},
  tch/.style={box, fill=cAha!10, text width=3.3cm, minimum height=0.9cm},
  tcho/.style={box, fill=cOurs!8, draw=cOurs!75, text width=3.3cm, minimum height=0.9cm},
  sig/.style={box, text width=3.1cm},
  arr/.style={->, line width=0.6pt, black},
  lab/.style={text=black, align=center},
]
\node[inner sep=0pt, draw=black, line width=0.4pt] (simg) at (1.2,4.5) {\includegraphics[width=2.0cm]{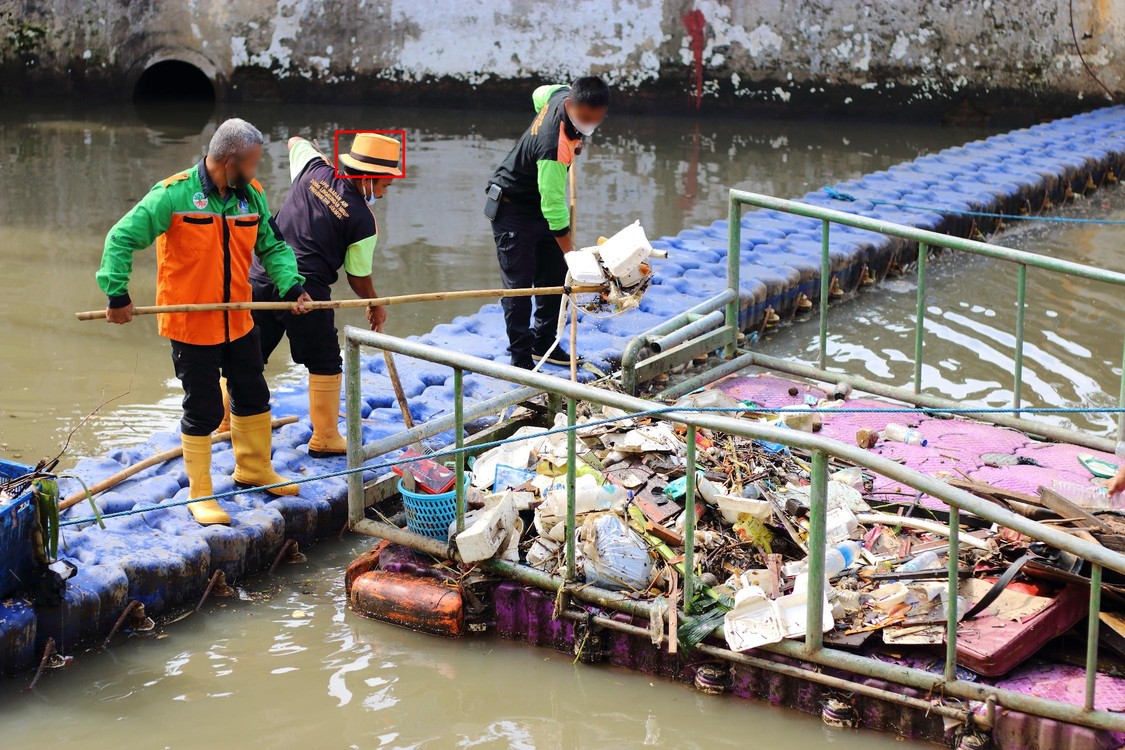}};
\draw[red, line width=0.6pt] ($(simg.north west)+(0.6018,-0.2364)$) rectangle ($(simg.north west)+(0.7138,-0.3120)$);
\node[lab, above=1pt of simg] {Student View $I$};
\node[stu] (stu)  at (4.0,4.5) {student $\pi_\theta$};
\node[box, lab, fill=white, font=\fontsize{10}{11.5}\selectfont, inner xsep=4pt, inner ysep=4pt, text width=4.85cm] (question) at (2.2,3.2)
      {\mbox{Question $q$: What is the primary}\\\mbox{color of the hat's crown?}};
\node[box, fill=orange!12, minimum width=3.5cm, minimum height=0.85cm] (roll) at (7.5,4.5)
      {rollout $y\sim\pi_\theta(\cdot\mid x_s)$\\[-1pt]{\itshape ``\dots the hat looks orange \dots\ (A)''}};
\node[box, minimum width=4.2cm, minimum height=0.7cm] (loss) at (16.4,4.5)
      {$\mathcal L_{\mathrm{OPD}}=\frac{1}{|y|}\sum_t\mathcal{D}\big(\pi_\theta(\cdot\mid x_s,y_{<t})\,\|\,q_t\big)$};
\draw[arr] (simg) -- (stu);
\draw[arr] (stu) -- (roll);
\draw[arr] (stu.south |- question.north) -- (stu.south);

\node[inner sep=0pt, draw=black, line width=0.4pt] (timg) at (1.2,1.2) {\includegraphics[height=1.3cm]{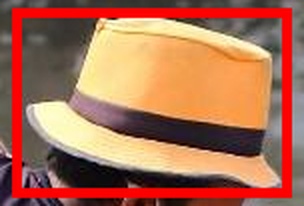}};
\node[lab, above=1pt of timg] {Privileged View $x_T$};
\node[tch]  (glance) at (6.9,3.0)  {\textbf{First Glance}\\$\theta_0(x_T)$, no prefix};
\node[tch]  (open)   at (6.9,1.85) {\textbf{Eyes Open} $p^{+}_t$\\$\theta_0(x_T,\,y_{<t})$};
\node[tcho] (closed) at (6.9,0.7)  {\textbf{Eyes Closed} $p^{\varnothing}_t$\\$\theta_0(\langle\mathrm{v}\rangle\langle/\mathrm{v}\rangle,\,q,\,y_{<t})$};
\draw[dashed, black, rounded corners=3pt, line width=0.5pt] (5.05,0.1) rectangle (8.75,3.62);
\node[lab, anchor=north] at (6.9,0.06) {frozen teacher $\theta_0$ (initial student)};

\draw[black, line width=0.6pt] (timg.east) -- (3.4,1.2) -- (3.4,2.5);
\draw[arr] (3.4,2.5) -- (4.9,2.5) |- (glance.west);
\draw[arr] (3.4,1.85) -- (open.west);
\draw[arr, densely dashed, cOurs!80] (3.4,1.85) |- (closed.west);

\node[sig] (gate) at (11.9,3.0) {\textbf{Conflict Gate}\\$\hat a(y)\neq\hat a_{\mathrm{glance}}$\,?};
\node[sig, fill=cOurs!6, draw=cOurs!65, minimum height=1.45cm] (evid) at (11.9,1.275)
      {\textbf{Privileged}\\\textbf{Visual Evidence}\\$r_t=\log p^{+}_t-\log p^{\varnothing}_t$};
\draw[arr] (glance.east) -- (gate.west);
\node[lab, inner sep=0.5pt, anchor=south west] at (9.08,3.05) {$\hat a_{\mathrm{glance}}$};
\draw[arr] (roll.east) -| (gate.north);
\node[lab, anchor=west] at (12.02,3.9) {answer $\hat a(y)$};
\draw[arr] (open.east)   -- (evid.west |- open.east);
\draw[arr] (closed.east) -- (evid.west |- closed.east);
\draw[black, line width=0.6pt, preaction={draw=white, line width=2.2pt}]
      (roll.south) -- (7.5,3.82) -- (9.0,3.82) -- (9.0,0.95);
\draw[arr] (9.0,2.1) -- ($(open.east)+(0,0.25)$);
\draw[arr] (9.0,0.95) -- ($(closed.east)+(0,0.25)$);
\fill[black] (9.0,2.1) circle[radius=0.025cm];
\node[lab, anchor=west] at (9.08,3.82) {prefix $y_{<t}$};

\node[box, fill=cOurs!12, draw=cOurs, line width=0.8pt, minimum width=4.3cm, minimum height=1.25cm] (tgt) at (16.4,2.15)
      {\textbf{Distillation Target}\\$q_t=\operatorname{softmax}\big(\log p^{+}_t+\beta_i\,r_t\big)$\\$\beta_i=\beta\,\gamma^{\mathbf{1}[\mathrm{conflict}]}$};
\coordinate (target-open-in) at ($(tgt.north west)+(0.9,0)$);
\coordinate (target-evidence-in) at ($(tgt.south west)+(0.9,0)$);
\draw[arr] (gate.east) -| (target-open-in);
\draw[arr] (evid.east) -| (target-evidence-in);
\draw[arr] (tgt.north) -- (loss.south);
\draw[arr, cOurs] (loss.north) -- (16.4,5.25) -- node[lab, above, text=cOurs!85!black, pos=0.5]{update $\theta$} (4.0,5.25) -- (stu.north);
\end{tikzpicture}

%% file: sec/5_experiments.tex
\section{Experiments}
\label{sec:exp}

\subsection{Experimental Setup}
\label{sec:setup}
\noindent\textbf{Model and Data.}
We train Qwen3.5-4B and 9B~\cite{qwen35} in non-thinking mode.
We use the \textbf{\textit{decontaminated}}\footnote{\label{fn:decontam}The released Vision-OPD training set has 6{,}241 questions. We found 21 of its images overlap with V$^*$ (11.0\% of the 191 V$^*$ images). We remove these 21 questions and train all methods, on the remaining 6{,}220.} Vision-OPD training pool~\cite{visionopd2026} of about 6.2K images with evidence boxes. We report results on random samples at different scales from 250 to the full 6.2K pool (\cref{fig:density}b).
All methods follow the Vision-OPD input protocol~\cite{visionopd2026}: the student receives the full image with the evidence box drawn on it and a one-sentence instruction to focus on that region, while the teacher receives a tight crop of the boxed region. 

\noindent\textbf{Implementation Details.}
We train for 38 steps with 96 prompts per step and 8 rollouts per prompt (temperature 1.0, at most 1{,}024 tokens).
We use a constant learning rate of $2\times10^{-6}$ after 5 warm-up steps and set $\beta=3.4$ and $\gamma=2$. We use JSD ($\alpha{=}0.5$) for the divergence $\mathcal{D}$ in \cref{eq:opd}. A 4B run takes under three hours on four B200 GPUs. The 9B run uses the same data, steps and hyperparameters (see \textbf{Appendix \textcolor{red}{I}} for more implementation details).

\noindent\textbf{Baselines.}
To compare methods under one protocol, we reproduce the recent OPSD methods with \textbf{\textit{complete}} public code: Vision-OPD~\cite{visionopd2026}, OPD-V~\cite{opdv2026}, CV-OPD~\cite{cvopd2026}, VAD~\cite{vad2026} and OPD-Aha~\cite{aha2026}, each with its official implementation.
All OPSD runs share the same student, the teacher crop, the rollout budget per step and the evaluation pipeline.
We also evaluate SFT and GRPO~\cite{shao2024deepseekmath} under the same setting. For other methods, \cref{tab:main} lists the numbers reported by their original papers in gray, for reference only.

\noindent\textbf{Evaluation.} 
Following the evaluation protocol of Vision-OPD~\cite{visionopd2026}, we report accuracy on V$^*$~\cite{wu2024vstar}, ZoomBench~\cite{zwz2026}, HR-Bench 4K/8K~\cite{wang2025hrbench} and MME-RealWorld EN/CN~\cite{zhang2025mmerealworld}.
Decoding is greedy with up to 32K tokens.
Answers are extracted by rules and a \textbf{\textit{corrected}}\footnote{The judge shared by prior work counts a response as correct when the first option letter in it matches the ground truth, even if the final answer differs~\cite{ead2026}; we fix it and report the corrected results (\textbf{Appendix \textcolor{red}{J}}).} gpt-oss-120B judge.

\begin{table*}[t]
\centering
\caption{Fine-grained perception. Accuracy (\%) on six benchmarks. \textcolor{black!55}{Gray}: results reported by the original papers or reproduced by others; \textbf{\textit{listed for reference only.}} Agentic models and ZwZ are built on older backbones ($^\ddagger$Qwen2.5-VL-7B, $^\S$Qwen3-VL-8B). \textbf{Bold} and \underline{underline} mark the best and second best within each re-evaluated block. Teacher: Self is a privileged self-teacher (a copy of the student on the zoomed crop); V-Zero instead distills from a Qwen3.5-27B teacher. \#Img: unique training images. $\Delta$: gain over the base model.}
\label{tab:main}
\setlength{\tabcolsep}{6pt}
\renewcommand{\arraystretch}{0.9}
\resizebox{\textwidth}{!}{%
\begin{tabular}{lllr|cc|cc|cc|cc}
\toprule
& & & & \multicolumn{2}{c|}{\textbf{Fine-Grained}} & \multicolumn{2}{c|}{\textbf{High-Resolution}} & \multicolumn{2}{c|}{\textbf{Real-World}} & & \\
\textbf{Method} & \textbf{Size} & \textbf{Teacher} & \textbf{\#Img} & V$^*$ & ZoomBench & HR-4K & HR-8K & MME-RW EN & MME-RW CN & \textbf{Avg.} & $\boldsymbol{\Delta}$ \\
\midrule
\grouprow{12}{Proprietary models (reported by~\cite{visionopd2026,ead2026}; also evaluated in~\cite{vad2026,cvopd2026,aha2026})}
\rep{GPT-5.2~\cite{gpt52_2025}} & \rep{--} & \rep{--} & \rep{--} & \rep{79.1} & \rep{50.9} & \rep{81.1} & \rep{78.4} & \rep{72.6} & \rep{68.8} & \rep{71.8} & \\
\rep{GPT-5.4~\cite{gpt54_2026}} & \rep{--} & \rep{--} & \rep{--} & \rep{77.0} & \rep{52.7} & \rep{84.0} & \rep{77.9} & \rep{74.2} & \rep{70.9} & \rep{72.8} & \\
\rep{Gemini 3.1 Pro~\cite{gemini31pro2026}} & \rep{--} & \rep{--} & \rep{--} & \rep{88.0} & \rep{61.2} & \rep{89.6} & \rep{86.9} & \rep{76.5} & \rep{73.3} & \rep{79.3} & \\
\rep{Gemini 3.5 Flash~\cite{gemini35flash2026}} & \rep{--} & \rep{--} & \rep{--} & \rep{89.0} & \rep{61.4} & \rep{89.1} & \rep{86.6} & \rep{75.3} & \rep{74.0} & \rep{79.2} & \\
\grouprow{12}{Large open-weight models (reported by~\cite{visionopd2026,ead2026}; also evaluated in~\cite{vad2026,cvopd2026,aha2026})}
\rep{GLM-4.6V~\cite{glm45v2025}} & \rep{106B} & \rep{--} & \rep{--} & \rep{86.9} & \rep{50.1} & \rep{82.1} & \rep{78.9} & \rep{65.6} & \rep{65.6} & \rep{71.5} & \\
\rep{Qwen3-VL-Instruct~\cite{qwen3vl}} & \rep{235B} & \rep{--} & \rep{--} & \rep{91.1} & \rep{56.1} & \rep{86.1} & \rep{80.4} & \rep{71.7} & \rep{69.0} & \rep{75.8} & \\
\rep{Qwen3.5~\cite{qwen35}} & \rep{397B} & \rep{--} & \rep{--} & \rep{88.0} & \rep{57.2} & \rep{89.4} & \rep{85.5} & \rep{74.8} & \rep{69.8} & \rep{77.4} & \\
\rep{Kimi-K2.6~\cite{kimik26_2026}} & \rep{1T} & \rep{--} & \rep{--} & \rep{88.5} & \rep{53.1} & \rep{81.9} & \rep{78.0} & \rep{69.2} & \rep{66.1} & \rep{72.8} & \\
\grouprow{12}{Thinking-with-images / inference-time methods (reported by~\cite{vad2026,aha2026}; also evaluated in~\cite{visionopd2026,ead2026,cvopd2026})}
\rep{DeepEyes~\cite{deepeyes2025}} & \rep{7B$^\ddagger$} & \rep{--} & \rep{--} & \rep{82.6} & \rep{46.2} & \rep{75.5} & \rep{70.0} & \rep{64.0} & \rep{62.4} & \rep{66.8} & \\
\rep{Thyme-RL~\cite{thyme2025}} & \rep{7B$^\ddagger$} & \rep{--} & \rep{--} & \rep{78.5} & \rep{46.2} & \rep{78.9} & \rep{70.9} & \rep{64.2} & \rep{61.4} & \rep{66.7} & \\
\rep{DeepEyesV2~\cite{deepeyesv2_2026}} & \rep{7B$^\ddagger$} & \rep{--} & \rep{--} & \rep{78.0} & \rep{46.1} & \rep{79.8} & \rep{72.6} & \rep{64.3} & \rep{61.9} & \rep{67.1} & \\
\rep{SenseNova-MARS~\cite{sensenovamars2026}} & \rep{8B$^\S$} & \rep{--} & \rep{--} & \rep{88.4} & \rep{49.0} & \rep{85.0} & \rep{77.3} & \rep{67.3} & \rep{65.7} & \rep{72.1} & \\
\grouprow{12}{Single forward-pass methods (reported by~\cite{aha2026,ead2026,visionopd2026}; also evaluated in~\cite{vad2026,cvopd2026})}
\rep{V-Zero~\cite{vzero2026}} & \rep{4B} & \rep{27B} & \rep{6.2K} & \rep{78.0} & \rep{56.5} & \rep{84.8} & \rep{80.9} & \rep{73.2} & \rep{71.1} & \rep{74.1} & \\
\rep{EAD~\cite{ead2026}} & \rep{4B} & \rep{Self} & \rep{6.2K} & \rep{91.6} & \rep{60.8} & \rep{86.4} & \rep{82.9} & \rep{72.4} & \rep{71.1} & \rep{77.5} & \\
\rep{EAD~\cite{ead2026}} & \rep{9B} & \rep{Self} & \rep{6.2K} & \rep{93.2} & \rep{62.0} & \rep{88.6} & \rep{84.6} & \rep{72.5} & \rep{71.1} & \rep{78.7} & \\
\rep{ZwZ~\cite{zwz2026}} & \rep{8B$^\S$} & \rep{--} & \rep{74K} & \rep{88.0} & \rep{56.7} & \rep{83.6} & \rep{81.8} & \rep{66.6} & \rep{68.1} & \rep{74.1} & \\
\midrule
\grouprow{12}{Qwen3.5-4B (reproduced by us with the decontaminated training set and the corrected judge protocol)}
Base model~\cite{qwen35} & 4B & -- & -- & 81.7 & 48.6 & 83.9 & 80.6 & 62.8 & 62.6 & 70.0 & -- \\
SFT & 4B & -- & 6.2K & 87.0 & 54.7 & 84.8 & 80.0 & 71.2 & 68.2 & 74.3 & \gain{+4.3} \\
GRPO~\cite{shao2024deepseekmath} & 4B & -- & 6.2K & 86.9 & 58.9 & 83.5 & 78.4 & 72.5 & 68.8 & 74.8 & \gain{+4.8} \\
Vision-OPD~\cite{visionopd2026} & 4B & Self & 6.2K & 89.5 & 59.3 & 82.5 & 79.4 & 71.3 & 68.6 & 75.1 & \gain{+5.1} \\
OPD-V~\cite{opdv2026} & 4B & Self & 6.2K & 89.0 & 59.9 & 84.4 & 78.5 & 70.6 & 69.4 & 75.3 & \gain{+5.3} \\
CV-OPD~\cite{cvopd2026} & 4B & Self & 6.2K & 90.1 & 60.4 & 83.3 & 80.5 & 72.3 & 70.2 & 76.1 & \gain{+6.1} \\
VAD~\cite{vad2026} & 4B & Self & 6.2K & \textbf{94.8} & 61.0 & 85.3 & 81.5 & 73.7 & 69.4 & 77.6 & \gain{+7.6} \\
OPD-Aha~\cite{aha2026} & 4B & Self & 6.2K & 91.6 & 59.3 & 87.4 & 83.3 & 73.9 & \underline{72.0} & 77.9 & \gain{+7.9} \\
\rowcolor{oursrow}\textbf{\method (ours)} & 4B & Self & \textcolor{red}{0.5K} & 92.7 & \underline{61.1} & \underline{88.5} & \underline{85.6} & \underline{74.8} & 71.8 & \underline{79.1} & \gain{\underline{+9.1}} \\
\rowcolor{oursrow}\textbf{\method (ours)} & 4B & Self & 6.2K & \underline{93.1} & \textbf{61.9} & \textbf{89.1} & \textbf{86.2} & \textbf{75.1} & \textbf{72.4} & \textbf{79.6} & \gain{\textbf{+9.6}} \\
\midrule
\grouprow{12}{Qwen3.5-9B (reproduced by us with the decontaminated training set and the corrected judge protocol)}
Base model~\cite{qwen35} & 9B & -- & -- & 85.9 & 52.8 & 84.9 & 80.1 & 70.3 & 67.2 & 73.5 & -- \\
GRPO~\cite{shao2024deepseekmath} & 9B & -- & 6.2K & 89.5 & 58.5 & 84.8 & 80.6 & 73.5 & 69.6 & 76.1 & \gain{+2.6} \\
SFT & 9B & -- & 6.2K & 90.1 & 58.8 & 84.9 & 81.6 & 73.3 & 68.5 & 76.2 & \gain{+2.7} \\
OPD-V~\cite{opdv2026} & 9B & Self & 6.2K & 88.5 & 61.5 & 85.5 & 83.3 & 71.0 & 69.1 & 76.5 & \gain{+3.0} \\
Vision-OPD~\cite{visionopd2026} & 9B & Self & 6.2K & 86.9 & \textbf{64.3} & 85.5 & 84.0 & 71.1 & 69.2 & 76.8 & \gain{+3.3} \\
CV-OPD~\cite{cvopd2026} & 9B & Self & 6.2K & 88.0 & 62.5 & 87.1 & 83.8 & 70.1 & 69.8 & 76.9 & \gain{+3.4} \\
VAD~\cite{vad2026} & 9B & Self & 6.2K & 86.9 & 63.3 & 86.5 & 85.4 & 72.8 & 70.3 & 77.5 & \gain{+4.0} \\
OPD-Aha~\cite{aha2026} & 9B & Self & 6.2K & 89.5 & 61.8 & 89.0 & 84.4 & 74.3 & \textbf{71.2} & 78.4 & \gain{+4.9} \\
\rowcolor{oursrow}\textbf{\method (ours)} & 9B & Self & \textcolor{red}{0.5K} & \underline{94.2} & 61.2 & \underline{90.4} & \underline{86.9} & \underline{74.4} & 70.4 & \underline{79.6} & \gain{\underline{+6.1}} \\
\rowcolor{oursrow}\textbf{\method (ours)} & 9B & Self & 6.2K & \textbf{94.7} & \underline{63.6} & \textbf{90.7} & \textbf{87.2} & \textbf{75.3} & \underline{70.6} & \textbf{80.4} & \gain{\textbf{+6.9}} \\
\bottomrule
\end{tabular}}
\vspace{-5pt}
\end{table*}

\begin{table*}[t]
\centering
\caption{Zero-shot transfer to ten held-out benchmarks. Perc., Hall.\ and Reas.: average accuracy on general perception, hallucination mitigation and multimodal reasoning. All: average over all 16 benchmarks, including the six fine-grained perception benchmarks of \cref{tab:main}. $\Delta$: change of All relative to the base model. \textbf{Bold} and \underline{underline}: best and second best within each model size.}
\label{tab:zeroshot}
\setlength{\tabcolsep}{2pt}
\renewcommand{\arraystretch}{1.15}
\resizebox{\textwidth}{!}{%
\begin{tabular}{l|ccc|cc|ccccc|cccc|c}
\toprule
& \multicolumn{3}{c|}{\textbf{General Perception}} & \multicolumn{2}{c|}{\textbf{Hallucination}} & \multicolumn{5}{c|}{\textbf{Multimodal Reasoning}} & \multicolumn{4}{c|}{\textbf{Average}} & \\
\textbf{Method} & MMStar & CV-Bench & MMVP & POPE & HallB. & MathVista & MathVerse & MATH-V & We-Math & DynaMath & Perc. & Hall. & Reas. & \textbf{All} & $\boldsymbol{\Delta}$ \\
\midrule
\grouprow{16}{Qwen3.5-4B}
Base model~\cite{qwen35} & 68.3 & 87.0 & 58.7 & \underline{89.5} & 76.1 & 82.0 & 74.2 & 63.9 & 58.8 & 57.0 & 71.3 & 82.8 & 67.2 & 71.0 & -- \\
Vision-OPD~\cite{visionopd2026} & 71.5 & 85.6 & 59.3 & 89.4 & 71.7 & 81.2 & 71.8 & 60.9 & 63.3 & 63.9 & 72.1 & 80.6 & 68.2 & 73.1 & \gain{+2.1} \\
OPD-V~\cite{opdv2026} & 71.4 & 85.8 & 58.7 & \underline{89.5} & 73.3 & 80.5 & 72.3 & 61.6 & 64.9 & 61.1 & 72.0 & 81.4 & 68.1 & 73.2 & \gain{+2.2} \\
CV-OPD~\cite{cvopd2026} & 71.5 & 86.6 & 56.7 & 89.3 & 71.8 & 81.0 & 72.4 & 63.5 & 67.1 & 57.8 & 71.6 & 80.6 & 68.4 & 73.4 & \gain{+2.4} \\
VAD~\cite{vad2026} & 73.5 & \textbf{88.2} & 58.0 & 89.2 & 72.7 & 83.1 & 73.4 & 64.5 & 69.9 & 65.5 & 73.2 & 81.0 & 71.3 & 75.2 & \gain{+4.2} \\
OPD-Aha~\cite{aha2026} & \underline{75.3} & 87.2 & \underline{61.3} & 89.0 & 74.5 & 82.6 & 77.1 & 69.6 & 75.5 & 65.6 & \underline{74.6} & 81.8 & 74.1 & 76.6 & \gain{+5.6} \\
\rowcolor{oursrow}\textbf{\method (6.2K)} & 74.3 & 87.0 & 60.5 & \textbf{89.6} & \textbf{77.4} & \textbf{84.1} & \textbf{79.3} & \textbf{71.2} & \textbf{76.9} & \textbf{66.3} & 73.9 & \textbf{83.5} & \textbf{75.6} & \underline{77.8} & \gain{\underline{+6.8}} \\
\rowcolor{oursrow}\textbf{\method (\textcolor{red}{0.5K})} & \textbf{75.6} & \underline{87.8} & \textbf{70.0} & \underline{89.5} & \underline{77.2} & \underline{83.9} & \underline{78.8} & \underline{70.8} & \underline{76.3} & \underline{65.8} & \textbf{77.8} & \underline{83.4} & \underline{75.1} & \textbf{78.1} & \gain{\textbf{+7.1}} \\
\midrule
\grouprow{16}{Qwen3.5-9B}
Base model~\cite{qwen35} & 67.9 & \underline{87.0} & 64.0 & 89.8 & 73.3 & 83.9 & 74.3 & 68.1 & 67.1 & 74.6 & 73.0 & 81.6 & 73.6 & 74.5 & -- \\
Vision-OPD~\cite{visionopd2026} & 74.7 & 82.8 & 66.0 & 89.9 & 73.6 & 82.8 & 74.1 & 68.4 & 74.1 & 76.2 & 74.5 & 81.8 & 75.1 & 76.5 & \gain{+2.0} \\
OPD-V~\cite{opdv2026} & 75.3 & 85.4 & 65.3 & 89.7 & 75.4 & 82.4 & 73.8 & 68.6 & 72.7 & 76.8 & 75.3 & 82.6 & 74.9 & 76.5 & \gain{+2.0} \\
CV-OPD~\cite{cvopd2026} & 75.7 & 85.5 & 66.7 & 89.8 & 74.0 & 82.2 & 76.6 & 68.5 & 72.2 & 76.7 & 76.0 & 81.9 & 75.2 & 76.8 & \gain{+2.3} \\
VAD~\cite{vad2026} & 77.0 & 84.7 & 68.7 & \underline{90.2} & \underline{76.4} & 83.6 & 75.0 & 69.2 & 73.1 & 77.0 & 76.8 & \underline{83.3} & 75.6 & 77.5 & \gain{+3.0} \\
OPD-Aha~\cite{aha2026} & \underline{77.5} & 86.3 & \underline{71.3} & \underline{90.2} & 74.4 & \underline{85.1} & \underline{77.4} & 72.7 & 76.8 & \underline{78.1} & \underline{78.4} & 82.3 & \underline{78.0} & 78.8 & \gain{+4.3} \\
\rowcolor{oursrow}\textbf{\method (\textcolor{red}{0.5K})} & \textbf{78.6} & \textbf{88.4} & \textbf{72.7} & 90.1 & 75.8 & \underline{85.1} & 75.0 & \underline{73.2} & \underline{77.5} & 78.0 & \textbf{79.9} & 83.0 & 77.8 & \underline{79.5} & \gain{\underline{+5.0}} \\
\rowcolor{oursrow}\textbf{\method (6.2K)} & 76.7 & 85.5 & 70.2 & \textbf{90.4} & \textbf{76.5} & \textbf{85.8} & \textbf{77.8} & \textbf{73.6} & \textbf{78.4} & \textbf{78.2} & 77.5 & \textbf{83.5} & \textbf{78.8} & \textbf{79.7} & \gain{\textbf{+5.2}} \\
\bottomrule
\end{tabular}}
\end{table*}

\subsection{Main Results for Fine-Grained Perception}
\label{sec:main}

As shown in \cref{tab:main}, \method reaches 79.1 average accuracy on Qwen3.5-4B with only 0.5K images, outperforming the base model (+9.1) and all reproduced OPSD baselines trained on $12\times$ more images: Vision-OPD (+4.0), OPD-V (+3.8), CV-OPD (+3.0), VAD (+1.5) and OPD-Aha (+1.2).
Training on the full 6.2K set further improves the average to 79.6.
The same trend holds for 9B: with only 0.5K images, \method reaches 79.6, outperforming the base model (+6.1) and all reproduced OPSD baselines trained on $12\times$ more images: OPD-V (+3.1), Vision-OPD (+2.8), CV-OPD (+2.7), VAD (+2.1) and OPD-Aha (+1.2).
Training on the full 6.2K set further improves the average to 80.4.

\noindent\textbf{Comparison with larger open-weight and proprietary models.}
\method also outperform larger open-weight models with up to $250\times$ more parameters, including Qwen3.5-397B (77.4), the strongest one in \cref{tab:main}. With only 4B parameters and 0.5K training images, \method already matches the leading proprietary models, including Gemini 3.1 Pro (79.3), the best one in \cref{tab:main}. Scaling to 9B parameters and the full 6.2K, \method further surpasses all reported and reproduced baselines. These gains are achieved despite training on a decontaminated dataset and evaluating with a corrected judge, both of which can only lower our scores relative to the setup from prior works.

\subsection{Zero-Shot Transfer to Other Visual Abilities}
\label{sec:zeroshot}
Post-training for fine-grained perception should not come at the cost of other visual abilities.
We therefore evaluate on three important visual abilities without any further training across ten held-out benchmarks (\cref{tab:zeroshot}): general perception (MMStar~\cite{chen2024mmstar}, CV-Bench~\cite{tong2024cambrian}, MMVP~\cite{tong2024mmvp}), hallucination (POPE~\cite{li2023pope}, HallusionBench~\cite{guan2024hallusionbench}) and multimodal reasoning (MathVista~\cite{lu2024mathvista}, MathVerse~\cite{zhang2024mathverse}, MATH-Vision~\cite{wang2024mathvision}, We-Math~\cite{qiao2025wemath}, DynaMath~\cite{zou2025dynamath}).
All models are evaluated with the official prompt of each benchmark, greedy decoding with thinking disabled and the same scorer (see \textbf{Appendix \textcolor{red}{K}} for the detailed experimental settings).

\noindent\textbf{General perception.}
The \method (0.5K) improves general perception the most at both sizes: reaching average scores of 77.8 (+6.5 over the base model) at 4B and 79.9 (+6.9) at 9B, ahead of the runner-up OPD-Aha in both (+3.2 and +1.5).
Fine-grained post-training corrects the \emph{myopia} of the base model, but it can overcorrect the model into the opposite extreme, \emph{hyperopia}.
This is most visible on CV-Bench at 9B, where every baseline falls below the base model (82.8--86.3 \vs 87.0), including the best method in fine-grained benchmarks: \method (6.2K).
We attribute this to their 6.2K training images, which overfit the model to fine details.
In contrast, \method (0.5K) is the only method that improves CV-Bench at both sizes (+0.8 at 4B and +1.4 at 9B), while also appearing strong on fine-grained perception in \cref{tab:main}.
This suggests that the key for OPSD to strike a balance between myopia and hyperopia is to raise the \textbf{\textit{efficiency of corrective supervision}}, rather than to simply train the model on ever more fine-grained images.

\noindent\textbf{Hallucination mitigation.}
At 4B, hallucination is the only ability for which all baselines underperform the base model (80.6--81.8 \vs 82.8 in the Hall. column), while \method (0.5K\&6.2K) is the only approach that achieves an improvement (83.4 and 83.5).
This can be attributed to the two flaws in existing distillation targets identified in \cref{sec:analysis}, both of which are themselves forms of hallucination.
Linguistic momentum (\cref{sec:capture}) causes the teacher to follow the student's prior words rather than the visual evidence, while image references in contrastive targets introduce a false belief (\cref{sec:refproblem}).
Although these flaws originate in the teacher's targets, they can propagate to the student through distillation.
\method removes both at the teacher level, breaking linguistic momentum without adding any false belief, and \textbf{\textit{the student inherits this behavior and hallucinates less.}}

\noindent\textbf{Multimodal Reasoning.}
\method (6.2K) achieves the best performance across all five reasoning benchmarks at both model sizes, improving the average from 67.2 to 75.6 (+8.4) at 4B and from 73.6 to 78.8 (+5.2) at 9B over the base model. We attribute these gains to its ability to break linguistic momentum. Multimodal reasoning produces long chains of thought. As shown in \cref{fig:capture}, the longer the student's prefix, the more closely the teacher follows it rather than the image. Among existing OPSD methods, \method is the most effective at breaking this linguistic momentum. By ``blinking'' at every token, its target keeps correcting the student even deep into a long chain, and the student learns to keep checking its reasoning against the image.

\noindent\textbf{\method achieves the strongest over four visual abilities.}
Averaged over all 16 benchmarks, the two settings of \method rank first and second at both sizes.
The 6.2K setting is best on fine-grained perception, hallucination mitigation and multimodal reasoning at both sizes, while the 0.5K setting is best on general perception. Linguistic momentum, the retention of privileged visual evidence and the efficiency of corrective supervision are general problems in OPSD: once they are addressed, the gains generalize to a broad range of visual understanding abilities.

\subsection{Corrective Supervision and Data Efficiency}
\label{sec:dataeff}
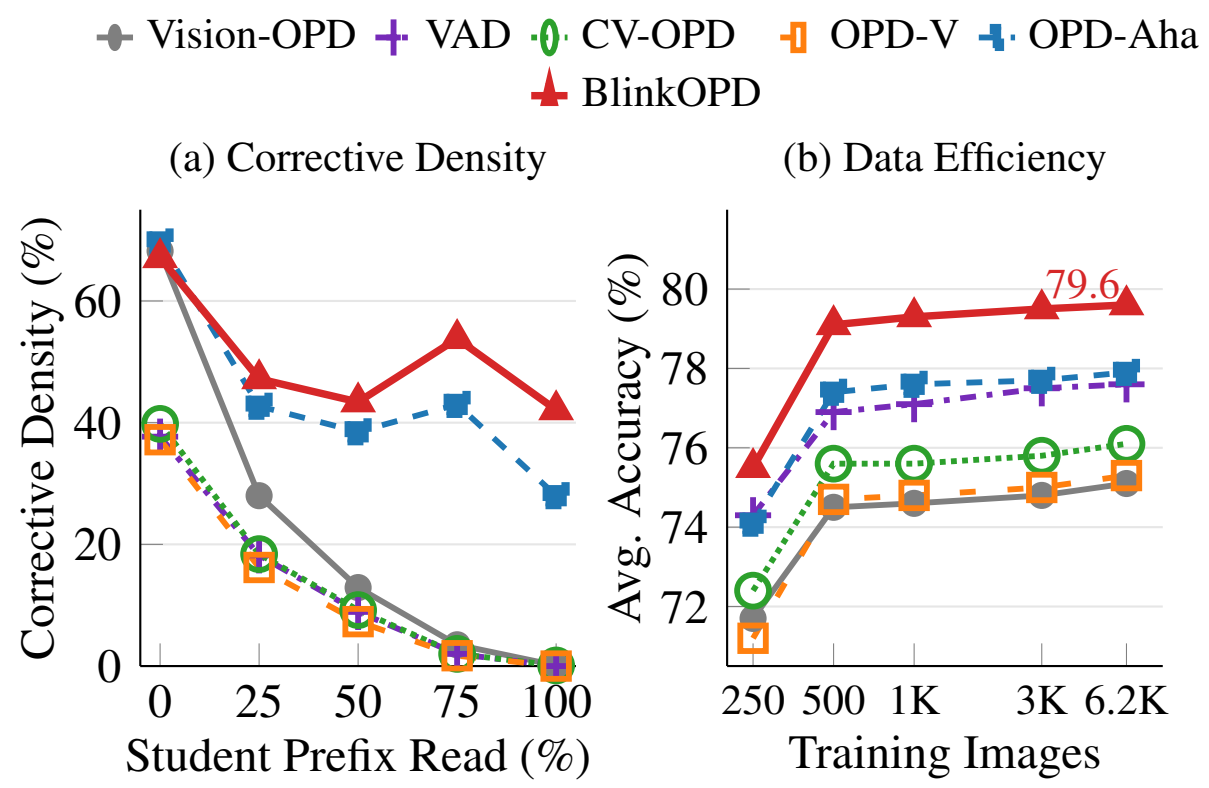
\begin{figure}[t]
\centering
\input{figs/dataeff}
\caption{Corrective supervision and data efficiency with Qwen3.5-4B.
(a) As the teacher reads more of the wrong prefix, the corrective density tends to decrease: Vision-OPD drops from 68\% to 0\%, VAD, CV-OPD and OPD-V from about 40\% to 0\%, and OPD-Aha declines to 28\%, whereas \method stays above 40\% throughout.
(b) Mean accuracy across the six fine-grained benchmarks at different training scales. \method leads at every scale and reaches 79.6\% with 6.2K images.}
\label{fig:density}
\end{figure}
To show that \method provides the densest corrective supervision, we define the corrective density of a distillation target as the share of student-wrong rollouts of the probe in \cref{sec:capture} on which it picks the correct option after a fraction $\tau$ of the student's own prefix has been read,
\begin{equation}
C(\tau)=\frac{1}{|\mathcal{W}|}\sum_{y\in\mathcal{W}}\mathbf{1}\Big[\argmax_{o\in\mathcal{O}_y}\,q^{\tau}_y(o)=g_y\Big],
\label{eq:density}
\end{equation}
with the notation of \cref{eq:follow,eq:margin}, where $g_y$ is the correct option of rollout $y$ and $\mathcal{O}_y$ is the set of its answer options.

\noindent\textbf{Why a few images can suffice.}
The higher corrective density of \method (\cref{fig:density}a) makes every rollout more useful for training: on a wrong rollout, more of the distillation tokens push the student toward the correct option instead of toward its own or another wrong answer.
\method also keeps the privileged visual evidence alive and resists linguistic momentum on long prefixes (\cref{fig:capture}), so its corrective signal does not fade too much as the student writes more.
Although \method (0.5K) uses $12\times$ fewer images than the baselines, each image and rollout yields far more useful supervision, so it can outperform the OPSD baselines.

\noindent\textbf{Data efficiency across training scales.}
\Cref{fig:density}b compares all six methods trained on 250, 500, 1K, 3K and 6.2K images.
\method leads at every scale.
With only 250 images, it reaches 75.5, 1.2 points above the strongest baseline at the same scale (74.3).
With 500 images, it reaches 79.1 and surpasses every baseline trained on the full 6.2K pool by at least 1.2 points, using $12\times$ fewer images.
Scaling to 6.2K further improves it to 79.6, 1.7 points above the strongest baseline at the same scale.
This shows the benefit of breaking linguistic momentum: by providing denser corrective supervision, \method makes better use of the training data at every scale, especially when the data are scarce, since more of its tokens correct the student.

\begin{table}[t]
\centering
\caption{Ablation\&sensitivity analysis of \method (0.5K) on 4B.}
\label{tab:ablation}
\setlength{\tabcolsep}{9.5pt}
\resizebox{\linewidth}{!}{%
\begin{tabular}{l|cc}
\toprule
Variant with Qwen3.5-4B as the base model & Avg. & $\Delta$ \\
\midrule
\rowcolor{oursrow}\method (0.5K) & 79.1 & -- \\
\midrule
\multicolumn{3}{l}{\textit{Ablation}} \\
\quad w/o first-glance gain (same $\beta$ for all rollouts) & 78.5 & \drop{$-0.6$} \\
\quad $\to$ gray-image reference & 77.4 & \drop{$-1.7$} \\
\quad $\to$ black-image reference & 75.4 & \drop{$-3.7$} \\
\quad $\to$ Gaussian-noise-image reference & 75.6 & \drop{$-3.5$} \\
\quad $\to$ mismatched-image reference & 76.4 & \drop{$-2.7$} \\
\quad $\to$ text-only reference & 78.2 & \drop{$-0.9$} \\
\quad w/o reference (plain OPSD, $q_t=\pT_t$) & 75.1 & \drop{$-4.0$} \\
\midrule
\multicolumn{3}{l}{\textit{Sensitivity}} \\
\quad JSD $\to$ forward KL & 77.6 & \drop{$-1.5$} \\
\quad JSD $\to$ reverse KL & 76.7 & \drop{$-2.4$} \\
\quad frozen $\theta_0$ teacher $\to$ EMA teacher & 76.1 & \drop{$-3.0$} \\
\bottomrule
\end{tabular}}
\end{table}

\subsection{Ablation and Sensitivity Analysis}
\label{sec:ablation}
We run all ablations with \method (0.5K) on Qwen3.5-4B.
In \cref{tab:ablation}, every component of \method contributes to its accuracy.
Removing the first-glance gain costs 0.6 points, and removing the reference altogether, which recovers plain OPSD, costs 4.0, the largest drop in the table.
Replacing the eyes-closed reference reproduces the ranking of \cref{fig:refmap}.
The text-only reference cancels the drift about as well as ours but injects a stronger false belief, and it comes closest ($-0.9$).
The image-based references leave about 20\% of the drift uncancelled and also inject false beliefs, and they fall further behind ($-1.7$ to $-3.7$).
Only the eyes-closed reference is low on both measures, and it performs best.
For the OPSD pipeline, JSD and a frozen teacher work best: replacing JSD by forward or reverse KL costs 1.5 and 2.4, and an EMA teacher costs 3.0, which suggests that a frozen privileged teacher provides the most stable target.

%% file: figs/dataeff.tex
\begin{tikzpicture}
\begin{groupplot}[
  group style={group size=2 by 1, horizontal sep=1.0cm},
  width=0.5355\linewidth, height=4.6cm,
  xlabel style={font=\small, yshift=3pt},
  ticklabel style={font=\small},
  ylabel style={font=\small, yshift=-5pt},
  title style={font=\footnotesize, yshift=-4pt},
  axis lines*=left, ymajorgrids, grid style={black!10},
  legend style={font=\footnotesize, draw=none, fill=none, inner sep=0.5pt, /tikz/every even column/.append style={column sep=1pt}},
  legend image post style={xscale=0.4},
  legend cell align=left,
  every axis plot/.append style={line width=1.1pt, mark size=2.0pt},
]
\nextgroupplot[
  title={(a) Corrective Density},
  xlabel={Student Prefix Read (\%)}, ylabel={Corrective Density (\%)},
  xmin=-5, xmax=105, xtick={0,25,50,75,100},
  ymin=0, ymax=75, ytick={0,20,40,60},
  legend to name=densityleg, legend columns=5,
]
\addplot[color=cVOPD, mark=*] coordinates {(0,68.2) (25,28.0) (50,12.9) (75,3.5) (100,0.1)};
\addlegendentry{Vision-OPD}
\addplot[color=cVAD, mark=+, mark size=3.4pt, dashdotted, mark options={solid, line width=1.1pt}] coordinates {(0,37.7) (25,18.2) (50,8.9) (75,2.0) (100,0.0)};
\addlegendentry{VAD}
\addplot[color=cCVOPD, mark=o, mark size=3.0pt, densely dotted, mark options={solid, fill=white}] coordinates {(0,39.8) (25,18.4) (50,9.2) (75,2.0) (100,0.1)};
\addlegendentry{CV-OPD}
\addplot[color=cOPDV, mark=square, mark size=2.4pt, loosely dashed, mark options={solid, fill=white}] coordinates {(0,37.2) (25,16.2) (50,7.3) (75,1.7) (100,0.0)};
\addlegendentry{OPD-V}
\addplot[color=cAha, mark=square*, dashed] coordinates {(0,69.7) (25,42.7) (50,38.5) (75,43.0) (100,28.0)};
\addlegendentry{OPD-Aha}
\addlegendimage{empty legend}\addlegendentry{}
\addlegendimage{empty legend}\addlegendentry{}
\addplot[color=cOurs, mark=triangle*, line width=1.4pt, mark size=2.5pt] coordinates {(0,67.1) (25,47.2) (50,43.4) (75,53.8) (100,42.1)};
\addlegendentry{\methodfig}
\nextgroupplot[
  title={(b) Data Efficiency},
  xlabel={Training Images}, ylabel={Avg. Accuracy (\%)},
  xmode=log, log basis x=10,
  xmin=200, xmax=8500, xtick={250,500,1000,3000,6220},
  xticklabels={250,500,1K,3K,6.2K},
  xticklabel style={font=\footnotesize},
  ymin=70.5, ymax=82, ytick={72,74,76,78,80},
]
\addplot[color=cVOPD, mark=*] coordinates {(250,71.7) (500,74.5) (1000,74.6) (3000,74.8) (6220,75.1)};
\addplot[color=cVAD, mark=+, mark size=3.4pt, dashdotted, mark options={solid, line width=1.1pt}] coordinates {(250,74.3) (500,76.9) (1000,77.1) (3000,77.5) (6220,77.6)};
\addplot[color=cCVOPD, mark=o, mark size=3.0pt, densely dotted, mark options={solid, fill=white}] coordinates {(250,72.4) (500,75.6) (1000,75.6) (3000,75.8) (6220,76.1)};
\addplot[color=cOPDV, mark=square, mark size=2.4pt, loosely dashed, mark options={solid, fill=white}] coordinates {(250,71.2) (500,74.7) (1000,74.8) (3000,75.0) (6220,75.3)};
\addplot[color=cAha, mark=square*, dashed] coordinates {(250,74.1) (500,77.4) (1000,77.6) (3000,77.7) (6220,77.9)};
\addplot[color=cOurs, mark=triangle*, line width=1.4pt, mark size=2.5pt] coordinates {(250,75.5) (500,79.1) (1000,79.3) (3000,79.5) (6220,79.6)};
\node[font=\footnotesize, text=cOurs, anchor=south east, inner sep=1pt] at (axis cs:6220,79.6) {79.6};
\end{groupplot}
\node[anchor=south] at ($(group c1r1.north)!0.5!(group c2r1.north)+(0,0.42cm)$) {\pgfplotslegendfromname{densityleg}};
\end{tikzpicture}

%% file: sec/6_conclusion.tex
\section{Conclusion}
\label{sec:conclusion}
We found that OPSD for fine-grained multimodal perception suffers from linguistic momentum and the privileged visual evidence decays along the response.
We further showed that existing contrastive references leave part of this drift uncancelled and inject false beliefs of their own.
We therefore proposed \method, which lets the teacher ``blink'' at every token of every rollout, and a first-glance gain strengthens the correction.
This simple change to the distillation target keeps the privileged evidence alive along the whole response and raises the corrective density of every rollout.
With only 0.5K images, \method outperforms all reproduced OPSD baselines trained on $12\times$ more images at both 4B and 9B; its 4B student already matches the strongest proprietary models, and its 9B student surpasses them.
Although \method is trained only on fine-grained perception data, we show that once these problems of OPSD are resolved in its distillation target, the student learns general habits that transfer well to broader visual abilities, such as general perception, hallucination mitigation and multimodal reasoning, without training on any of these tasks.

\noindent\textbf{Related work} and \textbf{Limitations} are discussed in \textbf{Appendix \textcolor{red}{L}} \& \textbf{\textcolor{red}{M}}.
\textbf{Qualitative examples} are in \textbf{Appendix \textcolor{red}{N}}.

%% file: sec/X_suppl.tex
\clearpage

\onecolumn

\begin{center}
    {\Large \textbf{Teach with Blinks: Breaking Linguistic Momentum in On-Policy Distillation for Fine-Grained Multimodal Perception}} \\[0.5cm]
    {\Large Supplementary Material}
\end{center}
\vspace{0.5cm}

\noindent\textbf{\Large  Overview}\\

\begin{enumerate}
    \item \textbf{Appendix \textcolor{red}{A}}: Experimental details of the analysis in \cref{sec:capture,sec:refproblem}.
    \item \textbf{Appendix \textcolor{red}{B}}: Control experiment: the content of the prefix drives linguistic momentum.
    \item \textbf{Appendix \textcolor{red}{C}}: Residual drift when only part of the prefix is read.
    \item \textbf{Appendix \textcolor{red}{D}}: Proof of Proposition 1: $\delta$ measures the residual drift in the contrast $r_t$.
    \item \textbf{Appendix \textcolor{red}{E}}: Implementing the eyes-closed reference in training.
    \item \textbf{Appendix \textcolor{red}{F}}: The trade-off of a global gain.
    \item \textbf{Appendix \textcolor{red}{G}}: Global gain versus first-glance gain.
    \item \textbf{Appendix \textcolor{red}{H}}: Training algorithm of \method.
    \item \textbf{Appendix \textcolor{red}{I}}: Training details.
    \item \textbf{Appendix \textcolor{red}{J}}: The corrected answer judge.
    \item \textbf{Appendix \textcolor{red}{K}}: Zero-shot transfer protocol.
    \item \textbf{Appendix \textcolor{red}{L}}: Related work.
    \item \textbf{Appendix \textcolor{red}{M}}: Limitations.
    \item \textbf{Appendix \textcolor{red}{N}}: Qualitative examples.\\
\end{enumerate}

\noindent\textbf{\Large \textcolor{red}{A}. Experimental details of the analysis}\\

\noindent\textbf{Probe set.}
We randomly sample 768 questions from the decontaminated Vision-OPD training pool~\cite{visionopd2026} of 6{,}220 questions (see Footnote~\ref{fn:decontam} of the main paper).
For each question, $\theta_0$ generates 4 student rollouts from the full image (temperature 1, at most 1{,}024 tokens), giving 3{,}072 rollouts, of which 1{,}732 are wrong and 1{,}340 are correct.
Each rollout is truncated at 0, 25, 50, 75 and 100\% of its tokens.
After each truncated prefix we append the answer cue \texttt{**Answer: (} and read the next-token probabilities over the option letters, normalized over the options, for the teacher (zoomed crop), for every reference and for every distillation target.
From these read-outs, we construct the distillation targets of Vision-OPD~\cite{visionopd2026}, VAD~\cite{vad2026}, CV-OPD~\cite{cvopd2026}, OPD-V~\cite{opdv2026} and OPD-Aha~\cite{aha2026} following the default setting of each original paper.
Note that this analysis uses the ground-truth answers to separate wrong from correct rollouts and to compute the follow rate and the survival score, whereas OPSD methods themselves require no ground-truth answers.

\noindent\textbf{Linguistic momentum (\cref{fig:capture}).}
Panel (a) reports the follow rate $F(\tau)$ of \cref{eq:follow}, where $\mathcal{W}$ contains the 1{,}732 student-wrong rollouts of the probe set.
\cref{tab:follow-full} lists the values corresponding to \cref{fig:capture}a.

\par\medskip\noindent\begin{minipage}{\linewidth}
\centering
\captionof{table}{Follow rate $F(\tau)$ (\%) on the 1{,}732 student-wrong rollouts, corresponding to \cref{fig:capture}a.}
\label{tab:follow-full}
\setlength{\tabcolsep}{6pt}
\small
\begin{tabular}{lccccc}
\toprule
Method & $\tau=0$ & 25\% & 50\% & 75\% & 100\% \\
\midrule
Vision-OPD~\cite{visionopd2026} & 23.2 & 66.7 & 83.5 & 95.8 & 99.8 \\
VAD~\cite{vad2026} & 51.4 & 75.2 & 87.5 & 97.1 & 100.0 \\
CV-OPD~\cite{cvopd2026} & 48.9 & 75.6 & 87.2 & 97.1 & 99.9 \\
OPD-V~\cite{opdv2026} & 50.3 & 78.2 & 89.2 & 97.6 & 100.0 \\
OPD-Aha~\cite{aha2026} & 19.4 & 46.9 & 49.2 & 43.8 & 64.3 \\
\rowcolor{oursrow}\method (ours) & 19.6 & 34.7 & 37.4 & 19.1 & 38.4 \\
\bottomrule
\end{tabular}
\end{minipage}\par\medskip

\noindent Panel (b) reports the survival score $S(\tau)$ of \cref{eq:margin}, where $\mathcal{K}\subseteq\mathcal{W}$ contains the 1{,}181 student-wrong rollouts on which the teacher picks the correct option before reading the prefix, \ie, the rollouts on which the zoomed crop holds the evidence that the student missed.
\cref{tab:survival-full} lists the values corresponding to \cref{fig:capture}b.

\par\medskip\noindent\begin{minipage}{\linewidth}
\centering
\captionof{table}{Survival score $S(\tau)$ on the 1{,}181 rollouts of $\mathcal{K}$, corresponding to \cref{fig:capture}b.}
\label{tab:survival-full}
\setlength{\tabcolsep}{6pt}
\small
\begin{tabular}{lccccc}
\toprule
Method & $\tau=0$ & 25\% & 50\% & 75\% & 100\% \\
\midrule
Vision-OPD~\cite{visionopd2026} & 1.60 & $-1.36$ & $-2.16$ & $-2.77$ & $-3.95$ \\
VAD~\cite{vad2026} & 0.15 & $-2.25$ & $-2.73$ & $-3.21$ & $-4.07$ \\
CV-OPD~\cite{cvopd2026} & 0.35 & $-2.17$ & $-2.78$ & $-3.30$ & $-4.18$ \\
OPD-V~\cite{opdv2026} & 0.05 & $-2.34$ & $-2.97$ & $-3.42$ & $-4.18$ \\
OPD-Aha~\cite{aha2026} & 8.92 & 3.14 & 1.51 & 1.43 & $-1.03$ \\
\rowcolor{oursrow}\method (ours) & 13.35 & 7.21 & 4.57 & 7.50 & 3.67 \\
\bottomrule
\end{tabular}
\end{minipage}\par\medskip

\noindent\textbf{Residual drift (\cref{fig:refmap}a).}
The gray, black and Gaussian-noise references replace the zoomed crop with an image of the same size filled with the mean color of the crop, with zeros, or with Gaussian noise.
The mismatched-image reference replaces it with the zoomed crop of another random probe question whose source image is different, resized to the same size, so that it contains real but unrelated visual content.
The residual drift $\delta$ (\cref{eq:kappa}) is computed on the 1{,}181 rollouts of $\mathcal{K}$ after the whole prefix.

\par\vskip 0pt plus 8\baselineskip\penalty-200\vskip 0pt plus -8\baselineskip
\vspace{\baselineskip}
\noindent\textbf{\Large \textcolor{red}{B}. Control: the content of the prefix drives linguistic momentum}\\

\begin{figure}[h]
\centering
\begin{tikzpicture}
\begin{axis}[
  width=0.6\linewidth, height=4.2cm,
  xmin=-5, xmax=105, xtick={0,25,50,75,100},
  ymin=0, ymax=100, ytick={0,20,40,60,80,100},
  xlabel={Student Prefix Read (\%)}, ylabel={Follow Rate (\%)},
  xlabel style={font=\small, yshift=3pt}, ylabel style={font=\small, yshift=-5pt},
  ticklabel style={font=\small},
  axis lines*=left, ymajorgrids, grid style={black!10},
  legend style={font=\footnotesize, at={(0.98,0.5)}, anchor=east, draw=none, fill=none},
  legend cell align=left,
  every axis plot/.append style={line width=1.1pt, mark size=2.0pt},
]
\draw[black!50, dashed, line width=0.5pt] (axis cs:-5,23.2) -- (axis cs:105,23.2);
\addplot[color=cVOPD, mark=*] coordinates {(0,23.2) (25,66.7) (50,83.5) (75,95.8) (100,99.8)};
\addlegendentry{Own prefix}
\addplot[color=cVOPD, mark=o, densely dashed, mark options={solid, fill=white}] coordinates {(0,22.9) (25,23.9) (50,25.3) (75,25.5) (100,24.4)};
\addlegendentry{Swapped prefix (same length)}
\end{axis}
\end{tikzpicture}
\caption{Follow rate of the raw teacher (Vision-OPD) on the 1{,}732 student-wrong rollouts when it reads the student's own prefix or a prefix of another question with the same length. The thin gray line marks the follow rate before reading any prefix (23.2\%).}
\label{fig:swap}
\end{figure}
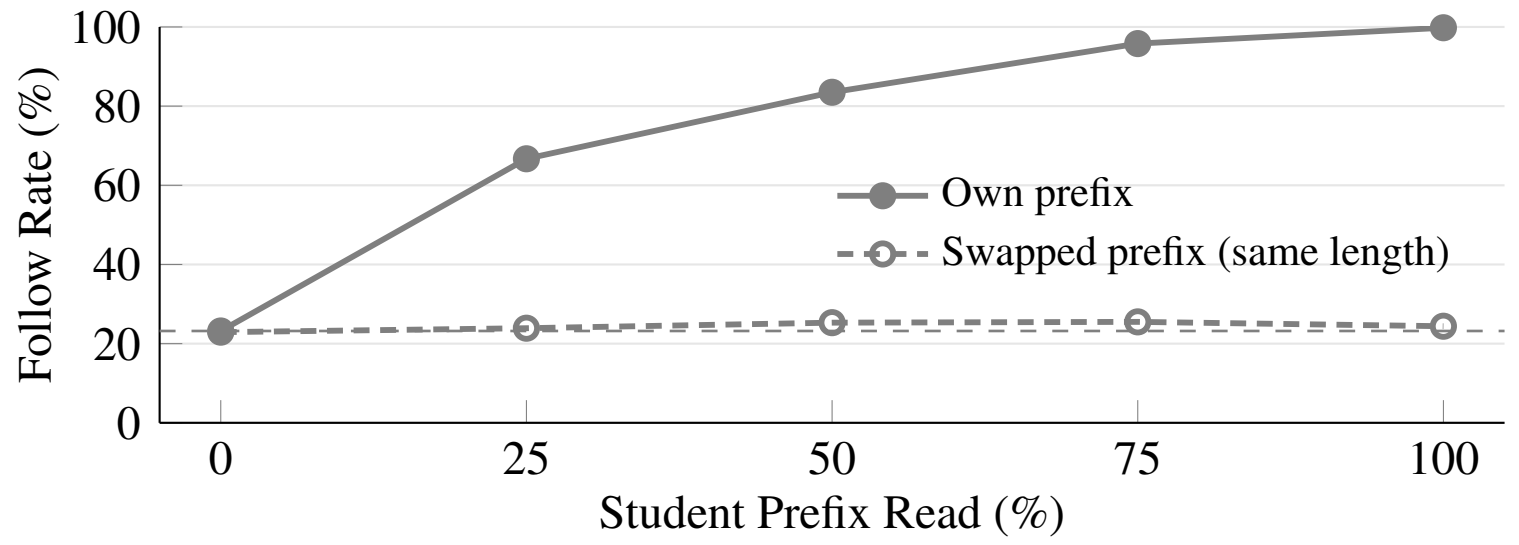

\noindent To check that linguistic momentum comes from the content of the prefix rather than its length, we repeat the probe of \cref{fig:capture}a on the same 1{,}732 student-wrong rollouts with the same teacher, but replace the student's prefix with a rollout of another question, taken from a different source image and truncated to the same number of tokens.
With its own prefix, the follow rate of the raw teacher (Vision-OPD) rises from 23.2\% to 99.8\%; with the swapped prefix, it stays between 22.9\% and 25.5\% (\cref{fig:swap}), \ie, at the level before reading any prefix.

\par\vskip 0pt plus 8\baselineskip\penalty-200\vskip 0pt plus -8\baselineskip
\vspace{\baselineskip}
\noindent\textbf{\Large \textcolor{red}{C}. Residual drift when only part of the prefix is read}\\

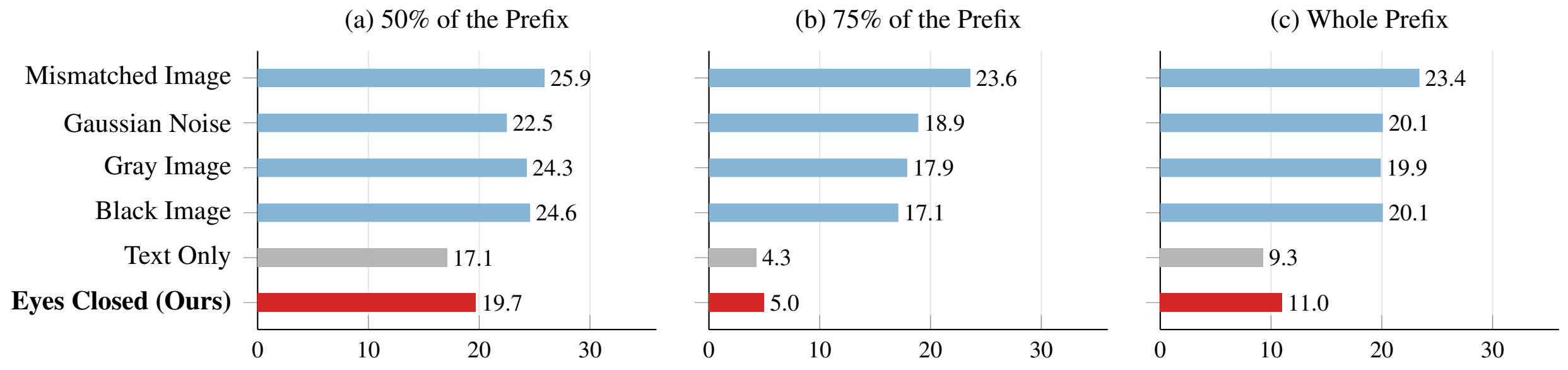
\begin{figure}[h]
\centering
\begin{tikzpicture}
\begin{groupplot}[
  group style={group size=3 by 1, horizontal sep=0.55cm, yticklabels at=edge left},
  width=0.33\linewidth, height=4.5cm,
  xbar, /pgf/bar width=5.5pt, /pgf/bar shift=0pt,
  y dir=reverse,
  ytick={1,2,3,4,5,6},
  yticklabels={Mismatched Image, Gaussian Noise, Gray Image, Black Image, Text Only, \textbf{Eyes Closed (Ours)}},
  yticklabel style={font=\footnotesize},
  ymin=0.4, ymax=6.6,
  xmin=0, xmax=36,
  xticklabel style={font=\scriptsize},
  title style={font=\footnotesize, yshift=-4pt},
  axis lines*=left, xmajorgrids, grid style={black!10},
  nodes near coords, nodes near coords style={font=\scriptsize, anchor=west, inner sep=1.5pt},
  every node near coord/.append style={/pgf/number format/.cd, fixed, fixed zerofill, precision=1},
  clip=false,
]
\nextgroupplot[title={(a) 50\% of the Prefix}]
\addplot[fill=cAha!55, draw=none] coordinates {(25.9,1) (22.5,2) (24.3,3) (24.6,4)};
\addplot[fill=cGray!45, draw=none] coordinates {(17.1,5)};
\addplot[fill=cOurs, draw=none] coordinates {(19.7,6)};
\nextgroupplot[title={(b) 75\% of the Prefix}]
\addplot[fill=cAha!55, draw=none] coordinates {(23.6,1) (18.9,2) (17.9,3) (17.1,4)};
\addplot[fill=cGray!45, draw=none] coordinates {(4.3,5)};
\addplot[fill=cOurs, draw=none] coordinates {(5.0,6)};
\nextgroupplot[title={(c) Whole Prefix}]
\addplot[fill=cAha!55, draw=none] coordinates {(23.4,1) (20.1,2) (19.9,3) (20.1,4)};
\addplot[fill=cGray!45, draw=none] coordinates {(9.3,5)};
\addplot[fill=cOurs, draw=none] coordinates {(11.0,6)};
\end{groupplot}
\end{tikzpicture}
\caption{Residual drift $\delta$ (\%, \cref{eq:kappa}) when the teacher and the reference read (a) 50\%, (b) 75\% or (c) all of the student prefix (\cref{fig:refmap}a).}
\label{fig:kappa-end}
\end{figure}

\noindent In \cref{fig:refmap}a, the residual drift $\delta$ (\cref{eq:kappa}) is measured between $\tau=0$ and the end of the whole prefix.
\cref{fig:kappa-end} shows $\delta$ when this end point is moved to $\tau=50\%$ and $75\%$, \ie, with $S^{\,\mathrm{end}}$ replaced by $S^{\,\tau}$ for both the teacher and the reference.

\par\vskip 0pt plus 8\baselineskip\penalty-200\vskip 0pt plus -8\baselineskip
\vspace{\baselineskip}
\noindent\textbf{\Large \textcolor{red}{D}. Proof of Proposition 1}\\

\noindent Here we show that the residual drift $\delta$ in \cref{eq:kappa}, although defined through the survival scores of the teacher and of the reference, directly measures how much of the prefix-induced drift remains in the contrast term $r_t$ of \cref{eq:tilt}.

\noindent\textbf{Notation.}
Fix a rollout $y\in\mathcal{K}$ and a truncation point $\tau$.
After reading the first $\tau$ of $y$, the teacher and the reference each assign a probability to every answer option $o\in\mathcal{O}_y$; we write them $p^{\tau}_{+,y}(o)$ and $p^{\tau}_{\mathrm{ref},y}(o)$.
These are $\pT_t$ and $\pR_t$ of \cref{eq:tilt} at the token position $t$ where the answer is read out, restricted to the answer options.
At this position, the contrast term $r_t$ of \cref{eq:tilt} is therefore a score for each option,
\begin{equation}
r^{\tau}_{y}(o)=\log p^{\tau}_{+,y}(o)-\log p^{\tau}_{\mathrm{ref},y}(o),\qquad o\in\mathcal{O}_y,
\label{eq:r-option}
\end{equation}
which is high for options that the teacher prefers more than the reference does.
($r_t$ in \cref{eq:tilt} carries no $\tau$ because it is defined at every token position $t$; here $t$ is the read-out position after the first $\tau$ of the prefix.)
As in \cref{eq:follow,eq:margin}, $g_y$ is the correct option of $y$ and $a_y$ is the wrong option chosen by the student.

\noindent\textbf{Proof of Proposition 1.}
\emph{Step 1.}
The survival scores of the teacher and of the reference are \cref{eq:margin} with $q^{\tau}_y$ replaced by $p^{\tau}_{+,y}$ and $p^{\tau}_{\mathrm{ref},y}$.
Subtracting them and regrouping the four log-probabilities by option gives, with \cref{eq:r-option},
\begin{equation}
\begin{aligned}
S_{+}(\tau)-S_{\mathrm{ref}}(\tau)
&=\frac{1}{|\mathcal{K}|}\sum_{y\in\mathcal{K}}\Big[\log\frac{p^{\tau}_{+,y}(g_y)}{p^{\tau}_{+,y}(a_y)}-\log\frac{p^{\tau}_{\mathrm{ref},y}(g_y)}{p^{\tau}_{\mathrm{ref},y}(a_y)}\Big]\\
&=\frac{1}{|\mathcal{K}|}\sum_{y\in\mathcal{K}}\Big[\underbrace{\big(\log p^{\tau}_{+,y}(g_y)-\log p^{\tau}_{\mathrm{ref},y}(g_y)\big)}_{r^{\tau}_y(g_y)}-\underbrace{\big(\log p^{\tau}_{+,y}(a_y)-\log p^{\tau}_{\mathrm{ref},y}(a_y)\big)}_{r^{\tau}_y(a_y)}\Big]\\
&=\frac{1}{|\mathcal{K}|}\sum_{y\in\mathcal{K}}\big[r^{\tau}_y(g_y)-r^{\tau}_y(a_y)\big]
\equiv S_{r}(\tau).
\end{aligned}
\label{eq:sr}
\end{equation}
$S_r(\tau)$ measures how strongly, on average over $\mathcal{K}$, the contrast term prefers the correct option to the student's option after the first $\tau$ of the prefix.

\emph{Step 2.}
Regrouping the four terms of $\Delta_{+}-\Delta_{\mathrm{ref}}$ by end point and applying \cref{eq:sr} at $\tau=0$ and at the end of the prefix,
\begin{equation}
\Delta_{+}-\Delta_{\mathrm{ref}}
=\big(S_{+}^{\,0}-S_{+}^{\,\mathrm{end}}\big)-\big(S_{\mathrm{ref}}^{\,0}-S_{\mathrm{ref}}^{\,\mathrm{end}}\big)
=\big(S_{+}^{\,0}-S_{\mathrm{ref}}^{\,0}\big)-\big(S_{+}^{\,\mathrm{end}}-S_{\mathrm{ref}}^{\,\mathrm{end}}\big)
=S_r^{\,0}-S_r^{\,\mathrm{end}}\equiv\Delta_r .
\label{eq:delta-regroup}
\end{equation}
$\Delta_r$ is how much reading the student prefix lowers the preference of $r_t$ for the correct option, \ie, the drift that remains in $r_t$.

\emph{Step 3.}
Dividing by $\Delta_{+}$ and using \cref{eq:kappa},
\begin{equation}
\delta=\frac{\Delta_{+}-\Delta_{\mathrm{ref}}}{\Delta_{+}}=\frac{\Delta_{r}}{\Delta_{+}} .
\end{equation}
Hence $\delta$ is the drift that remains in $r_t$, as a share of the teacher's own drift: $\delta=0$ means that the student prefix does not move $r_t$ at all, and any $\delta>0$ is drift that $r_t$ still carries and that the gain $\beta$ in \cref{eq:tilt} amplifies.

\noindent\hfill$\square$

\par\vskip 0pt plus 8\baselineskip\penalty-200\vskip 0pt plus -8\baselineskip
\vspace{\baselineskip}
\noindent\textbf{\Large \textcolor{red}{E}. Implementing the eyes-closed reference in training}\\

\noindent For efficiency, our trainer (verl~\citesupp{verl2025}) does not pad sequences of different lengths to a common length; it packs several sequences end to end into one long row and marks where each sequence starts. This appendix describes how we build the eyes-closed input in this packed format, and two details that matter for Qwen3.5.

\noindent\textbf{Building the eyes-closed input.}
For every sample, the trainer stores the token ids, the attention mask, a four-row position tensor (one text row used for packing and causal masking, and the three M-RoPE rows $t,h,w$~\citesupp{qwen2vl}), and the index where the response begins.
To build the eyes-closed input, we
\begin{enumerate}
    \item drop every visual-content token (the image-pad tokens) but keep the vision start/end markers;
    \item keep the three M-RoPE rows of all remaining tokens at their full-input values;
    \item renumber the text row contiguously, since it only marks where each sequence starts and which tokens may attend to which;
    \item shift the response start by the number of removed tokens; and
    \item pass no pixel values, so the vision encoder is skipped.
\end{enumerate}
On average, 414 tokens are removed per training row.

\noindent\textbf{Deleting rather than masking.}
We delete the image-content tokens instead of masking them in attention.
In decoders that also mix tokens through recurrent states or convolutions, such as the gated delta-rule layers~\citesupp{gateddeltanet2025} of Qwen3.5, masking attention alone would leave these paths from the image open; deleting the tokens removes every path from the image to the remaining tokens.
Keeping the original positions leaves the relative positions among the remaining tokens, and hence all position-dependent attention logits among them, unchanged.

\noindent\textbf{Resetting the linear-attention state.}
The gated delta-rule layers carry a recurrent state from token to token, so in a packed row the state of one sequence would leak into the next; it must be reset at each sequence boundary.
Without the reset, the KL between the trainer's and the rollout engine's token distributions has a median of 0.12, whereas with the reset it drops to 0.005.
We apply the reset to every forward pass of student, teacher and reference.

\par\vskip 0pt plus 8\baselineskip\penalty-200\vskip 0pt plus -8\baselineskip
\vspace{\baselineskip}
\noindent\textbf{\Large \textcolor{red}{F}. The trade-off of a global gain}\\

\noindent \cref{sec:glance} states that a single global gain does not suit all rollouts: a large gain recovers the correct option on student-wrong rollouts but pushes the distillation target away from it on student-correct ones.
Here we formulate this trade-off and then run an experiment on the probe of \cref{sec:capture} to show that it exists.

\noindent\textbf{Formulation.}
A global gain gives the same gain $\beta$ to every rollout, \ie, it is the gain of \cref{eq:beta} with $\gamma=1$.
On each rollout $y$ of the probe, the teacher and the eyes-closed reference read a fraction $\tau$ of the prefix, and we form the distillation target $q^{\tau}_y$ of \cref{eq:target} with this gain.
On a student-wrong rollout, the target should recover the correct option $g_y$; we measure this on the student-wrong rollouts $\mathcal{W}$ by the corrective density $C(\tau)$ of \cref{eq:density}.
On a student-correct rollout, the target should keep $g_y$; we measure this on the student-correct rollouts $\overline{\mathcal{W}}$ by the \emph{retention}
\begin{equation}
R(\tau)=\frac{1}{|\overline{\mathcal{W}}|}\sum_{y\in\overline{\mathcal{W}}}\mathbf{1}\Big[\argmax_{o\in\mathcal{O}_y}\,q^{\tau}_y(o)=g_y\Big],
\label{eq:retention}
\end{equation}
\ie, the share of student-correct rollouts on which the target still picks the correct option.
Let $C(\beta)$ and $R(\beta)$ denote the two measures under the global gain $\beta$, averaged over the read-out points $\tau$; a good gain needs both a high $C(\beta)$ and a high $R(\beta)$.
The two are in trade-off when $C(\beta)$ increases and $R(\beta)$ decreases as $\beta$ grows: then no single $\beta$ maximizes both, and every gain that corrects more student-wrong rollouts keeps fewer student-correct ones.

\noindent\textbf{Setup.}
The probe contains 768 training questions with four $\theta_0$ rollouts each, \ie, 1{,}732 student-wrong rollouts $\mathcal{W}$ and 1{,}340 student-correct rollouts $\overline{\mathcal{W}}$ (\textbf{Appendix \textcolor{red}{A}}).
We sweep the global gain $\beta$ from 0, where the target is the teacher alone, to 12 in steps of 1, and for each $\beta$ average $C(\tau)$ and $R(\tau)$ over $\tau\in\{25,50,75,100\}\%$.

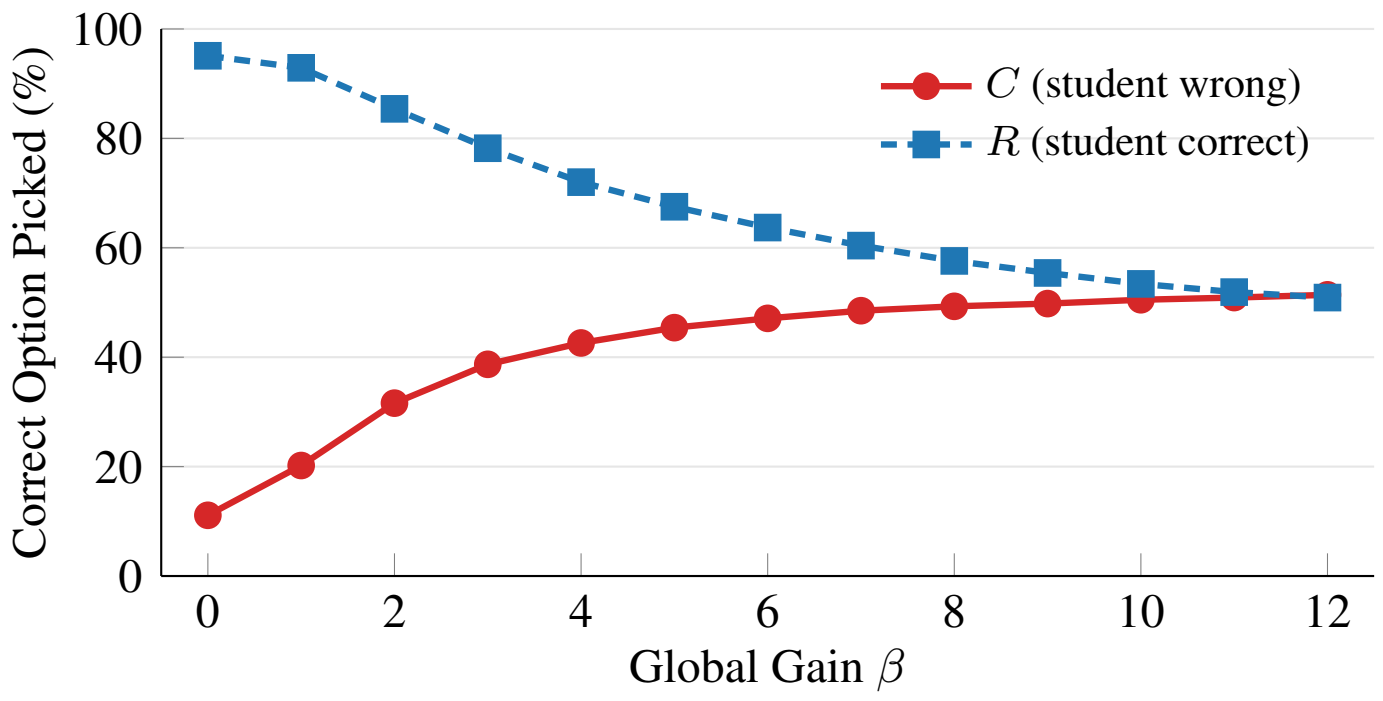
\begin{figure}[h]
\centering
\begin{tikzpicture}
\begin{axis}[
  width=0.55\linewidth, height=5.2cm,
  xmin=-0.5, xmax=12.5, xtick={0,2,4,6,8,10,12},
  ymin=0, ymax=100, ytick={0,20,40,60,80,100},
  xlabel={Global Gain $\beta$}, ylabel={Correct Option Picked (\%)},
  xlabel style={font=\small, yshift=3pt}, ylabel style={font=\small, yshift=-4pt},
  ticklabel style={font=\small},
  axis lines*=left, ymajorgrids, grid style={black!10},
  legend style={font=\footnotesize, draw=none, fill=none, at={(0.97,0.97)}, anchor=north east},
  legend cell align=left,
  every axis plot/.append style={line width=1.2pt, mark size=2.0pt},
]
\addplot[color=cOurs, mark=*] coordinates {(0,11.1) (1,20.2) (2,31.6) (3,38.7) (4,42.6) (5,45.4) (6,47.1) (7,48.5) (8,49.3) (9,49.8) (10,50.5) (11,50.9) (12,51.4)};
\addlegendentry{$C$ (student wrong)}
\addplot[color=cAha, mark=square*, densely dashed, mark options={solid}] coordinates {(0,95.1) (1,92.9) (2,85.4) (3,78.2) (4,72.0) (5,67.5) (6,63.7) (7,60.4) (8,57.6) (9,55.4) (10,53.4) (11,51.9) (12,50.9)};
\addlegendentry{$R$ (student correct)}
\end{axis}
\end{tikzpicture}
\caption{Trade-off of a global gain on the probe of \cref{sec:capture}. Each point is the share of rollouts on which the target picks the correct option, averaged over the read-out points $\tau=25$, 50, 75 and 100\%. Raising the global gain $\beta$ raises the corrective density $C$ on student-wrong rollouts but lowers the retention $R$ on student-correct ones.}
\label{fig:tradeoff}
\end{figure}

\noindent\textbf{The trade-off exists.}
As $\beta$ grows, the corrective density rises and the retention falls at every gain we test (\cref{fig:tradeoff}).
No global gain therefore gives both a high corrective density and a high retention: a gain that is large enough to correct the student-wrong rollouts damages the student-correct ones.
The gain should thus be raised only on rollouts that are likely to be wrong, which is what the first-glance gain of \cref{sec:glance} does (\textbf{Appendix \textcolor{red}{G}}).

\par\vskip 0pt plus 8\baselineskip\penalty-200\vskip 0pt plus -8\baselineskip
\vspace{\baselineskip}
\noindent\textbf{\Large \textcolor{red}{G}. Global gain versus first-glance gain}\\

\noindent\textbf{Experimental setup.}
We compare the global gain with the first-glance gain on the probe of \textbf{Appendix \textcolor{red}{F}}, using the corrective density $C$ and retention $R$ defined there.
We fix $\gamma=2$, the factor used in training, and sweep the base gain $\beta$ from 0.1 to 12 in steps of 0.1, comparing with the global gain of the same $\beta$.
As in training, the first glance is the teacher's top option on the crop before it reads any of the prefix ($\tau=0$).
The glance needs no label: it only compares the student's answer with the teacher's own answer on the crop, which the frozen teacher computes once per question before training.

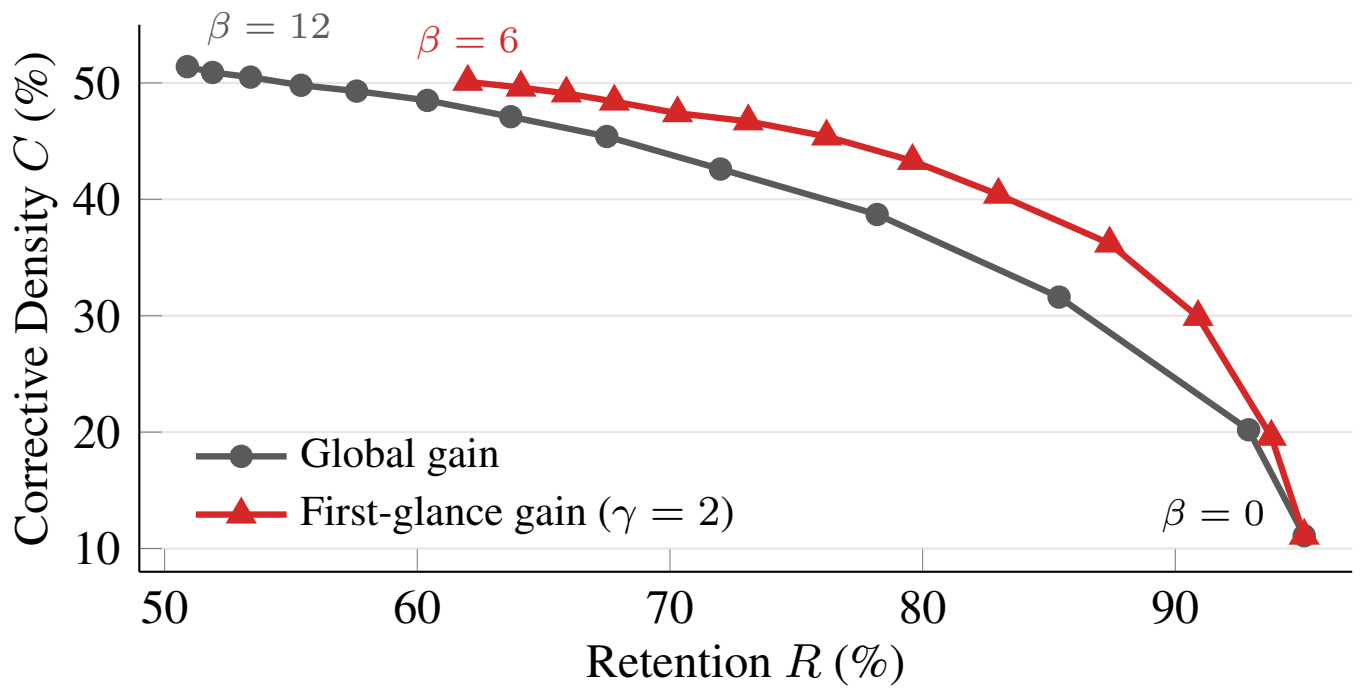
\begin{figure}[h]
\centering
\begin{tikzpicture}
\begin{axis}[
  width=0.55\linewidth, height=5.2cm,
  xmin=49, xmax=97, xtick={50,60,70,80,90},
  ymin=8, ymax=55, ytick={10,20,30,40,50},
  xlabel={Retention $R$ (\%)}, ylabel={Corrective Density $C$ (\%)},
  xlabel style={font=\small, yshift=3pt}, ylabel style={font=\small, yshift=-4pt},
  ticklabel style={font=\small},
  axis lines*=left, ymajorgrids, grid style={black!10},
  legend style={font=\footnotesize, draw=none, fill=none, at={(0.03,0.03)}, anchor=south west},
  legend cell align=left, clip=false,
  every axis plot/.append style={line width=1.2pt, mark size=2.0pt},
]
\addplot[color=cGray, mark=*, mark size=1.6pt] coordinates {(95.1,11.1) (92.9,20.2) (85.4,31.6) (78.2,38.7) (72.0,42.6) (67.5,45.4) (63.7,47.1) (60.4,48.5) (57.6,49.3) (55.4,49.8) (53.4,50.5) (51.9,50.9) (50.9,51.4)};
\addlegendentry{Global gain}
\addplot[color=cOurs, mark=triangle*, mark size=2.2pt] coordinates {(95.1,11.1) (93.8,19.6) (90.9,29.9) (87.4,36.2) (83.0,40.4) (79.6,43.3) (76.2,45.4) (73.1,46.7) (70.3,47.4) (67.8,48.4) (65.9,49.1) (64.1,49.6) (62.0,50.1)};
\addlegendentry{First-glance gain ($\gamma=2$)}
\node[font=\scriptsize, anchor=east] at (axis cs:94.3,12.8) {$\beta=0$};
\node[font=\scriptsize, anchor=south west, color=cGray] at (axis cs:50.9,51.6) {$\beta=12$};
\node[font=\scriptsize, anchor=south, color=cOurs] at (axis cs:62.0,50.4) {$\beta=6$};
\end{axis}
\end{tikzpicture}
\caption{First-glance gain ($\gamma=2$) against the global gain on the probe of \cref{sec:capture}: $C$ against $R$, each averaged over the four read-out points, as $\beta$ is swept from 0 to 12 for the global gain and from 0 to 6 for the first-glance gain, so that both reach a largest gain of 12. The first-glance gain lies above the global gain.}
\label{fig:glance}
\end{figure}

\noindent\textbf{The first-glance gain dominates the global gain.}
\cref{fig:glance} compares the two gains over the corrective-density--retention trade-off of \textbf{Appendix \textcolor{red}{F}}.
At the same retention, the first-glance gain corrects more student-wrong rollouts than the global gain; at the same corrective density, it keeps more student-correct rollouts.
In summary, over the whole range covered by both curves, the curve of the first-glance gain lies above and to the right of that of the global gain, \ie, the first-glance gain dominates the global gain on both measures.
This agrees with the ablation in \cref{tab:ablation}, where using a global gain instead lowers the average accuracy on the six fine-grained benchmarks by 0.6 points.

\par\vskip 0pt plus 8\baselineskip\penalty-200\vskip 0pt plus -8\baselineskip
\vspace{\baselineskip}
\noindent\textbf{\Large \textcolor{red}{H}. Training algorithm of \method}\\

\noindent \cref{alg:blink} summarizes one training step of \method, combining the eyes-closed reference (\cref{sec:eyes}), the first-glance gain (\cref{sec:glance}) and the distillation target (\cref{sec:objective}). The corresponding framework is shown in \cref{fig:overview} of the main paper.

\begin{algorithm}[ht]
\caption{One training step of \method}
\label{alg:blink}
\begin{algorithmic}[1]
\Require student $\pi_\theta$; teacher (a copy of the student) $\theta_0$; base gain $\beta$; factor $\gamma$; first glances $\hat a^{\text{glance}}(x)$ precomputed with $\theta_0$ on $x_T$
\For{each prompt $x=(I,I^+,q)$ in the batch}
  \State sample rollouts $y_1,\dots,y_n\sim\pi_\theta(\cdot\mid x_s)$
  \For{each rollout $y_i$}
    \State $\beta_i\gets\beta\,\gamma^{\,\scalebox{0.85}{$\mathbf{1}[\hat a(y_i)\neq\hat a^{\text{glance}}(x)]$}}$ \Comment{\cref{eq:beta}}
    \State $\pT_t\gets p_{\theta_0}(\cdot\mid x_T,y_{i,<t})$ for all $t$ \Comment{eyes open}
    \State $\pC_t\gets p_{\theta_0}(\cdot\mid \langle\text{v}\rangle\langle/\text{v}\rangle,q,y_{i,<t})$ \Comment{eyes closed}
    \State $q_{i,t}\gets\softmax\big(\log\pT_t+\beta_i[\log\pT_t-\log\pC_t]\big)$
  \EndFor
\EndFor
\State update $\theta$ by minimizing \cref{eq:opd} with $q_t=q_{i,t}$
\end{algorithmic}
\end{algorithm}

\par\vskip 0pt plus 8\baselineskip\penalty-200\vskip 0pt plus -8\baselineskip
\vspace{\baselineskip}
\noindent\begin{minipage}{\linewidth}
\noindent\textbf{\Large \textcolor{red}{I}. Training details}\\

\centering
\captionof{table}{Hyperparameters of \method (Qwen3.5-4B; 9B identical).}
\label{tab:hparams}
\setlength{\tabcolsep}{4pt}
\small
\begin{tabular}{ll}
\toprule
Item & Value \\
\midrule
Training pool & Vision-OPD, decontaminated (6.2K) \\
Unique images $K$ & 500 (random) \\
Student input & full image + box + focus instruction \\
Teacher input & tight crop around the box \\
Teacher & frozen $\theta_0$ \\
Batch / rollouts per prompt & 96 prompts / 8 \\
Rollout temperature, top-$p$ & 1.0, 1.0 \\
Max prompt / response length & 8{,}192 / 1{,}024 tokens \\
Optimizer & AdamW, lr $2\times10^{-6}$, 5 warm-up steps, constant \\
Steps & 38 \\
Divergence & JSD, $\alpha=0.5$, student top-100 + tail \\
Gain & $\beta=3.4$, $\gamma=2$ \\
Seed & 42 \\
Hardware & 4--8 NVIDIA B200 \\
\bottomrule
\end{tabular}
\end{minipage}\par

\par\vskip 0pt plus 8\baselineskip\penalty-200\vskip 0pt plus -8\baselineskip
\vspace{\baselineskip}
\noindent\textbf{\Large \textcolor{red}{J}. The corrected answer judge}\\

\noindent The answer judge shared by prior work has an issue: it marks a response as correct when the first option-like letter in it matches the ground truth, even if the final answer is a different option.
For example, for a response ending in ``\dots\ so the sign is blue. Answer: (C)'' with ground truth A, the judge takes the ``A'' of the word ``Answer'' as the chosen option and counts the response as correct.
Almost all such false positives have ground truth A and occur on MME-RealWorld, and they inflate average scores by 0.3--2.1 points.
EAD~\cite{ead2026} also identified this issue and switched to a Qwen3.5-122B judge.
For a fair comparison with Vision-OPD and the baselines reproduced from it, we keep their judge, gpt-oss-120B, and only fix this rule: the corrected judge uses only the parsed final answer.
We re-score all models with it, including the reproduced baselines, and report the corrected numbers.
We do not compare with EAD, since its code had not been released when we submitted this paper, so it could neither be reproduced under our protocol nor re-scored with the corrected judge.

\par\vskip 0pt plus 8\baselineskip\penalty-200\vskip 0pt plus -8\baselineskip
\vspace{\baselineskip}
\noindent\textbf{\Large \textcolor{red}{K}. Zero-shot transfer protocol}\\

\noindent\textbf{Data and prompts.}
All ten benchmarks are taken from the official releases at pinned revisions: MMStar (1{,}500), CV-Bench (2{,}638), MMVP (300), POPE (COCO random/popular/adversarial), HallusionBench, MathVista testmini (1{,}000), MathVerse testmini (3{,}940, five versions), MATH-Vision test (3{,}040), We-Math testmini (1{,}740) and DynaMath (5{,}010 = 501 seeds $\times$ 10 variants).
Each question uses the verbatim prompt of the benchmark's official evaluation code or paper.
We keep the official MMVP ground truth, which contains two swapped pairs that later versions correct.
Decoding is greedy with thinking disabled, as in the fine-grained evaluation.
CV-Bench is reported as $0.5\times\big((\text{ADE20K}+\text{COCO})/2+\text{Omni3D}\big)$.

\noindent\textbf{Scoring.}
CV-Bench, MMStar and MMVP: we strip markdown and extract the option from the answer marker (``Answer:'', ``the answer is'', \texttt{\textbackslash boxed\{\}}, or a leading option); if no option can be extracted, GPT-OSS-120B matches the response to an option with the official VLMEvalKit prompt, and unmatched responses count as wrong.
MathVista and MathVerse: official rules, with the judge used only to extract answers that the rules miss (MathVista) or with the official two-step extract-and-compare prompt for free-form answers (MathVerse).
MATH-Vision, We-Math and DynaMath: official rule-based scorers.
POPE and HallusionBench: rule-based yes/no extraction.
The judge runs at temperature 0 with a shared cache, so every model is scored by the same verdicts.
It is called for about 38\% of MathVerse responses, as intended by its official protocol, and for 0--12\% elsewhere; re-scoring the same responses changes averages by at most about 0.15.

\noindent\textbf{Reported metrics.}
We report the strict score on We-Math, the average accuracy over all ten variants on DynaMath, and question accuracy on HallusionBench; MATH-Vision is evaluated on its full test set.

\par\vskip 0pt plus 8\baselineskip\penalty-200\vskip 0pt plus -8\baselineskip
\vspace{\baselineskip}

\input{sec/2_related}

\noindent \cref{tab:positioning} compares \method with the five OPSD methods that we reproduce in \cref{sec:setup}.
All of them distill a privileged teacher on the zoomed crop into a student on the full image, and all of them run the student alone in a single pass at test time.

\begin{table}[h]
\centering
\caption{Comparison with the reproduced OPSD baselines. Granularity: the level at which the strength of the reference varies. Labels: whether ground-truth answers are used during training ($^*$CV-OPD skips the questions on which no rollout is correct).}
\label{tab:positioning}
\setlength{\tabcolsep}{3pt}
\small
\begin{tabular}{>{\raggedright\arraybackslash}p{2.2cm}>{\raggedright\arraybackslash}p{1.0cm}>{\raggedright\arraybackslash}p{3.4cm}>{\raggedright\arraybackslash}p{4.0cm}>{\raggedright\arraybackslash}p{2.3cm}>{\raggedright\arraybackslash}p{1.5cm}>{\raggedright\arraybackslash}p{0.9cm}}
\toprule
Method & Teacher & Reference & Role of the reference & Gain & Granularity & Labels \\
\midrule
Vision-OPD~\cite{visionopd2026} & EMA & none & -- & -- & -- & no \\
OPD-V~\cite{opdv2026} & EMA & crop with a random black patch & on/off weight on each token's loss & -- & per token & no \\
CV-OPD~\cite{cvopd2026} & EMA & another region of the same image & hinge term that keeps the student closer to the teacher & fixed weight & per token & yes$^*$ \\
VAD~\cite{vad2026} & EMA & pixelated crop & projects the teacher--student shift onto the evidence direction & capped shift & per token & no \\
OPD-Aha~\cite{aha2026} & frozen & gray (mean-color) image & contrast inside the distillation target & fixed $\beta$ & global & no \\
\rowcolor{oursrow}\method (ours) & frozen & \textbf{eyes closed}: image-content tokens removed, frame kept & contrast inside the distillation target & raised on first-glance conflicts & per rollout & no \\
\bottomrule
\end{tabular}
\end{table}

\par\vskip 0pt plus 8\baselineskip\penalty-200\vskip 0pt plus -8\baselineskip
\vspace{\baselineskip}
\noindent\textbf{\Large \textcolor{red}{M}. Limitations}\\

\noindent \method is general and can be applied to any multimodal model family.
However, owing to limited compute, our experiments cover only the Qwen3.5 family at 4B and 9B, the model familiy and sizes on which the OPSD methods we compare with are mainly evaluated.
We leave the validation on other model families to future work.

Although the eyes-closed reference is the only reference that is low on both residual drift and false belief (\cref{fig:refmap}), part of the drift is still left uncancelled, and the probability of the false belief remains slightly above that obtained with the real image.
Likewise, once the teacher reads the student prefix, \method has the lowest follow rate and the highest corrective density of all distillation targets (\cref{fig:capture,fig:density}), yet its target still follows the wrong prefix on some rollouts, and its corrective density stays below the level reached before any prefix is read.
Linguistic momentum is therefore reduced but not fully removed, and we leave references and targets that close this remaining gap to future work.

\par\vskip 0pt plus 8\baselineskip\penalty-200\vskip 0pt plus -8\baselineskip
\vspace{\baselineskip}
\noindent\textbf{\Large \textcolor{red}{N}. Qualitative examples}\\

\begin{figure*}[p]
\centering
\noindent\begin{tikzpicture}[baseline=(img.center)]
\node[anchor=south west, inner sep=0pt] (img) at (0,0) {\includegraphics[height=3.0cm]{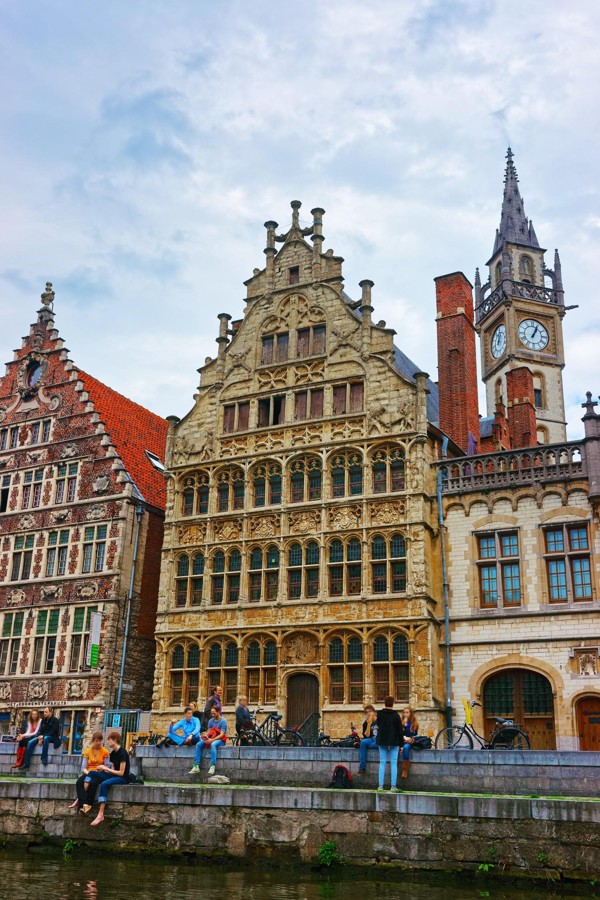}};
\begin{scope}[x={(img.south east)}, y={(img.north west)}]
\draw[red, line width=1pt] (0.8133,0.5911) rectangle (0.9600,0.6711);
\end{scope}
\end{tikzpicture}\hspace{0.25cm}%
\begin{tikzpicture}[baseline=(z.center)]
\node[inner sep=0pt, draw=red, line width=1pt] (z) {\includegraphics[height=3.0cm]{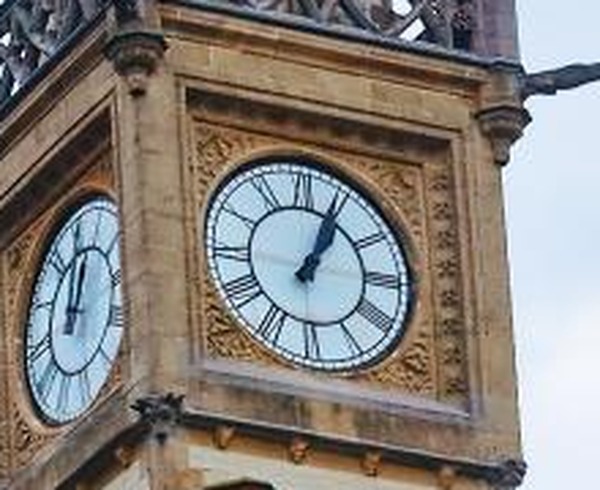}};
\end{tikzpicture}\hfill
\begin{minipage}[c]{0.40\textwidth}\small
\textbf{(a) ZoomBench.} What time is displayed on the clock face on the right side of the tower?\\[3pt]
A.~12:45 \quad B.~1:20 \quad C.~12:58 \quad D.~1:04\\[3pt]
\textbf{Ground truth:} D.~1:04
\end{minipage}

\vspace{4pt}
{\small
\begin{tabular*}{\textwidth}{@{\extracolsep{\fill}}lccccccc}
\toprule
Model & Base & Vision-OPD & OPD-V & CV-OPD & VAD & OPD-Aha & \cellcolor{oursrow}\textbf{\method (ours)} \\
\midrule
Answer & B~\textcolor{BrickRed}{\xmark} & B~\textcolor{BrickRed}{\xmark} & B~\textcolor{BrickRed}{\xmark} & B~\textcolor{BrickRed}{\xmark} & B~\textcolor{BrickRed}{\xmark} & C~\textcolor{BrickRed}{\xmark} & \cellcolor{oursrow}\textbf{D}~\textcolor{ForestGreen}{\cmark} \\
\bottomrule
\end{tabular*}}

\vspace{10pt}
\noindent\begin{tikzpicture}[baseline=(img.center)]
\node[anchor=south west, inner sep=0pt] (img) at (0,0) {\includegraphics[height=3.0cm]{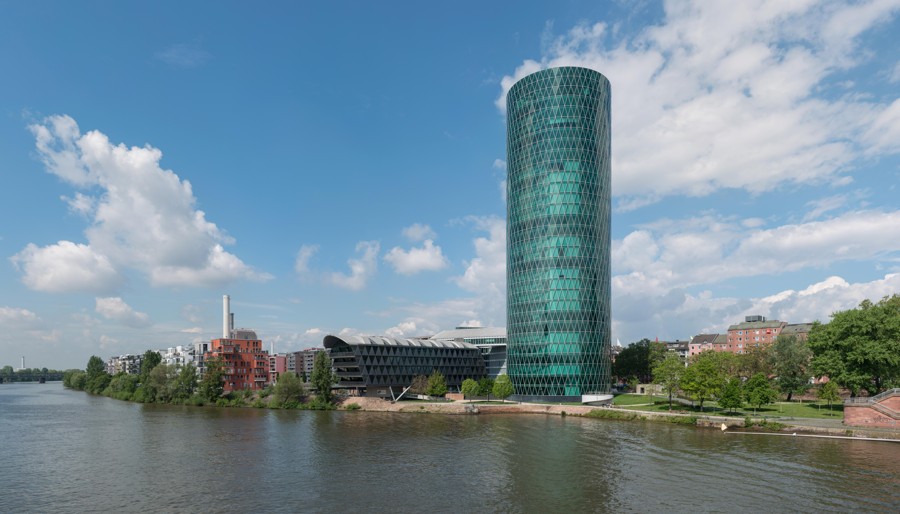}};
\begin{scope}[x={(img.south east)}, y={(img.north west)}]
\draw[red, line width=1pt] (0.8307,0.3254) rectangle (0.8776,0.3847);
\end{scope}
\end{tikzpicture}\hspace{0.25cm}%
\begin{tikzpicture}[baseline=(z.center)]
\node[inner sep=0pt, draw=red, line width=1pt] (z) {\includegraphics[height=3.0cm]{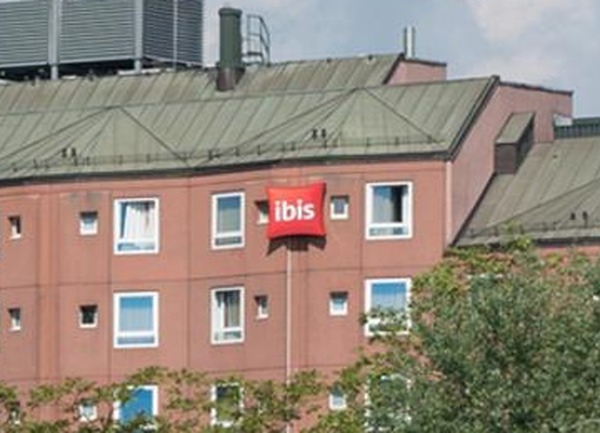}};
\end{tikzpicture}\hfill
\begin{minipage}[c]{0.40\textwidth}\small
\textbf{(b) HR-Bench 8K.} Tell me the shape of the signboard attached to the building?\\[3pt]
A.~Oval \quad B.~Rectangular\\[3pt]
C.~Triangular \quad D.~Circular\\[3pt]
\textbf{Ground truth:} B.~Rectangular
\end{minipage}

\vspace{4pt}
{\small
\begin{tabular*}{\textwidth}{@{\extracolsep{\fill}}lccccccc}
\toprule
Model & Base & Vision-OPD & OPD-V & CV-OPD & VAD & OPD-Aha & \cellcolor{oursrow}\textbf{\method (ours)} \\
\midrule
Answer & C~\textcolor{BrickRed}{\xmark} & A~\textcolor{BrickRed}{\xmark} & D~\textcolor{BrickRed}{\xmark} & C~\textcolor{BrickRed}{\xmark} & D~\textcolor{BrickRed}{\xmark} & D~\textcolor{BrickRed}{\xmark} & \cellcolor{oursrow}\textbf{B}~\textcolor{ForestGreen}{\cmark} \\
\bottomrule
\end{tabular*}}

\vspace{10pt}
\noindent\begin{tikzpicture}[baseline=(img.center)]
\node[anchor=south west, inner sep=0pt] (img) at (0,0) {\includegraphics[height=3.0cm]{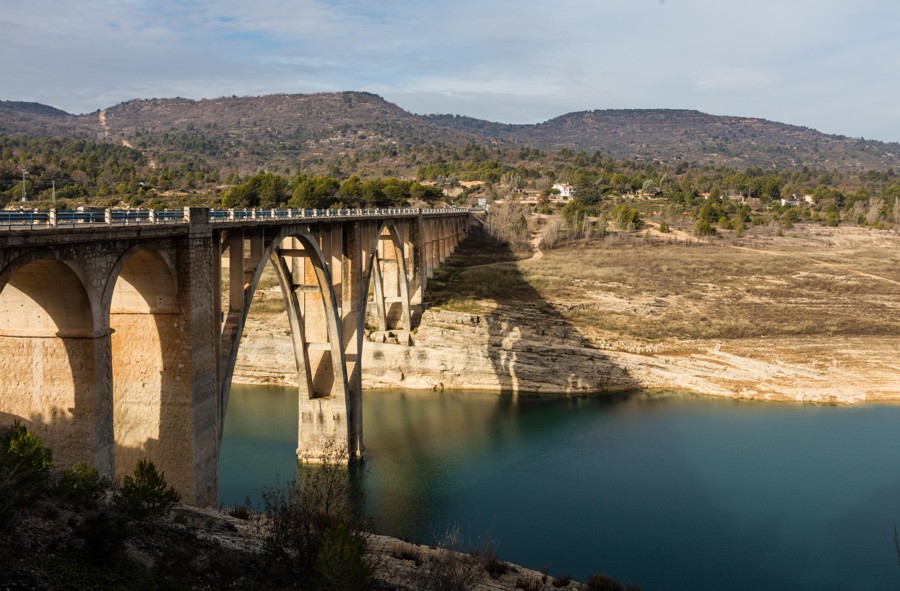}};
\begin{scope}[x={(img.south east)}, y={(img.north west)}]
\draw[red, line width=1pt] (0.5221,0.6413) rectangle (0.5495,0.6730);
\end{scope}
\end{tikzpicture}\hspace{0.25cm}%
\begin{tikzpicture}[baseline=(z.center)]
\node[inner sep=0pt, draw=red, line width=1pt] (z) {\includegraphics[height=3.0cm]{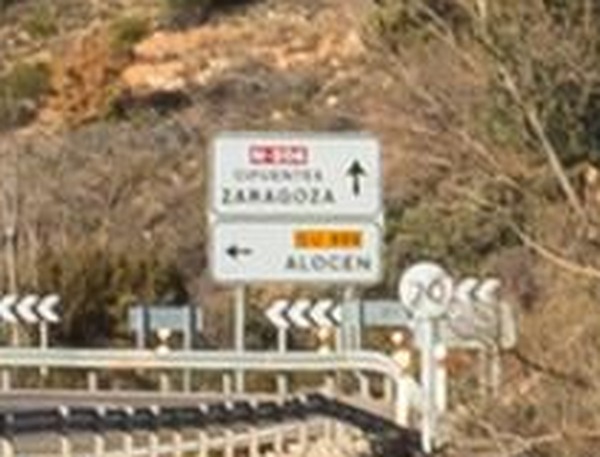}};
\end{tikzpicture}\hfill
\begin{minipage}[c]{0.40\textwidth}\small
\textbf{(c) HR-Bench 8K.} What is the direction indicated on the sign pointing towards Alocén?\\[3pt]
A.~Straight \quad B.~Left \quad C.~Right \quad D.~Downward\\[3pt]
\textbf{Ground truth:} B.~Left
\end{minipage}

\vspace{4pt}
{\small
\begin{tabular*}{\textwidth}{@{\extracolsep{\fill}}lccccccc}
\toprule
Model & Base & Vision-OPD & OPD-V & CV-OPD & VAD & OPD-Aha & \cellcolor{oursrow}\textbf{\method (ours)} \\
\midrule
Answer & A~\textcolor{BrickRed}{\xmark} & C~\textcolor{BrickRed}{\xmark} & C~\textcolor{BrickRed}{\xmark} & C~\textcolor{BrickRed}{\xmark} & C~\textcolor{BrickRed}{\xmark} & C~\textcolor{BrickRed}{\xmark} & \cellcolor{oursrow}\textbf{B}~\textcolor{ForestGreen}{\cmark} \\
\bottomrule
\end{tabular*}}

\vspace{10pt}
\noindent\begin{tikzpicture}[baseline=(img.center)]
\node[anchor=south west, inner sep=0pt] (img) at (0,0) {\includegraphics[height=3.0cm]{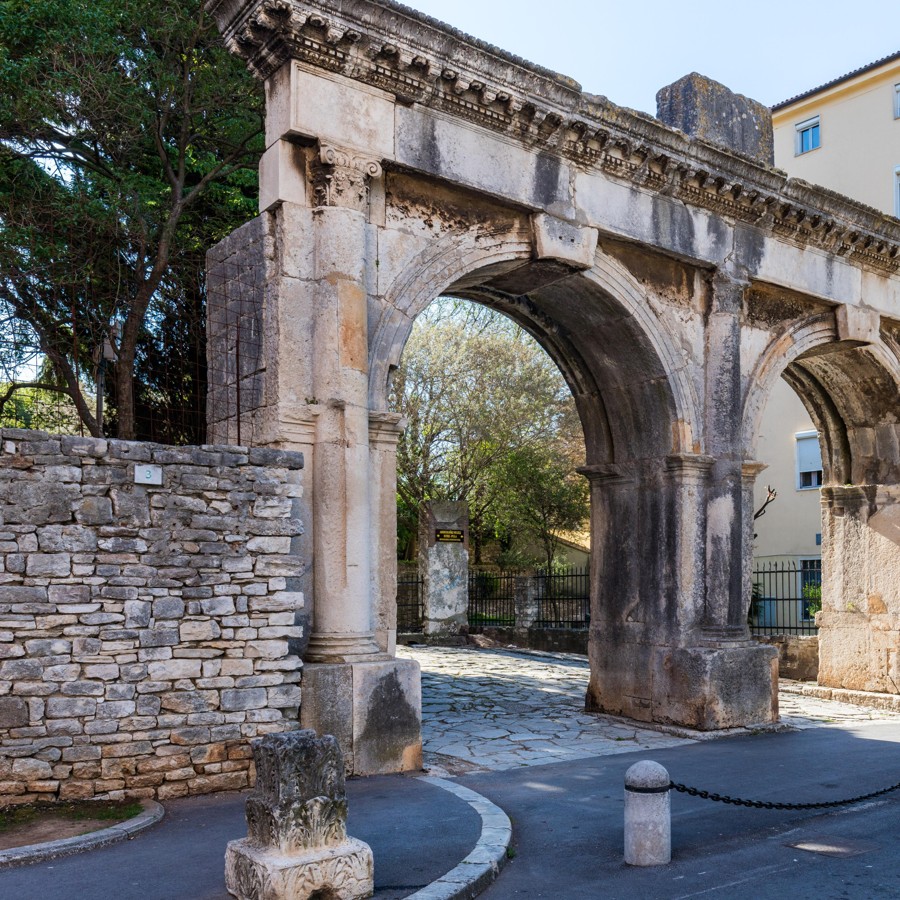}};
\begin{scope}[x={(img.south east)}, y={(img.north west)}]
\draw[red, line width=1pt] (0.4625,0.3750) rectangle (0.5375,0.4437);
\end{scope}
\end{tikzpicture}\hspace{0.25cm}%
\begin{tikzpicture}[baseline=(z.center)]
\node[inner sep=0pt, draw=red, line width=1pt] (z) {\includegraphics[height=3.0cm]{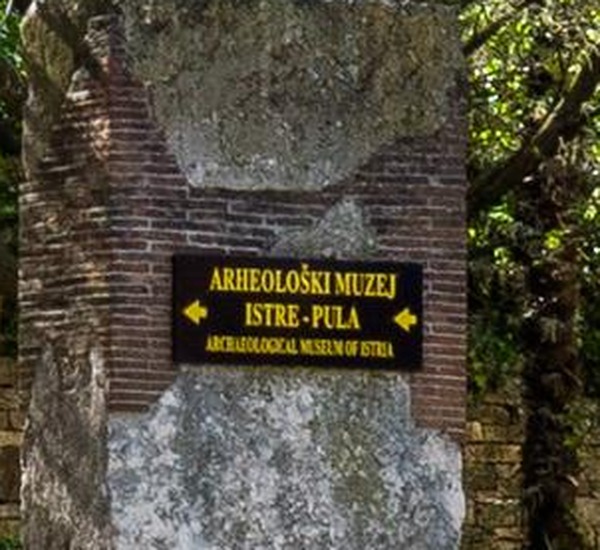}};
\end{tikzpicture}\hfill
\begin{minipage}[c]{0.40\textwidth}\small
\textbf{(d) HR-Bench 4K.} What direction arrows are shown on the black sign?\\[3pt]
A.~Two right arrows \quad B.~No arrows\\[3pt]
C.~Two left arrows \quad D.~Both left and right arrows\\[3pt]
\textbf{Ground truth:} C.~Two left arrows
\end{minipage}

\vspace{4pt}
{\small
\begin{tabular*}{\textwidth}{@{\extracolsep{\fill}}lccccccc}
\toprule
Model & Base & Vision-OPD & OPD-V & CV-OPD & VAD & OPD-Aha & \cellcolor{oursrow}\textbf{\method (ours)} \\
\midrule
Answer & D~\textcolor{BrickRed}{\xmark} & D~\textcolor{BrickRed}{\xmark} & D~\textcolor{BrickRed}{\xmark} & D~\textcolor{BrickRed}{\xmark} & D~\textcolor{BrickRed}{\xmark} & D~\textcolor{BrickRed}{\xmark} & \cellcolor{oursrow}\textbf{C}~\textcolor{ForestGreen}{\cmark} \\
\bottomrule
\end{tabular*}}

\caption{Qualitative examples of fine-grained perception. Left: the test image; the red box marks the region that the question refers to and is drawn only for this figure, since all models see the original image. Middle: the boxed region, enlarged. Right: the question, its options and the ground truth. Bottom: the answer of each model (\textcolor{ForestGreen}{\cmark}~correct, \textcolor{BrickRed}{\xmark}~wrong).}
\label{fig:qual}
\end{figure*}

\begin{figure*}[p]
\centering
\noindent\begin{minipage}[c]{0.40\textwidth}\centering
\includegraphics[height=3.0cm, width=\linewidth, keepaspectratio]{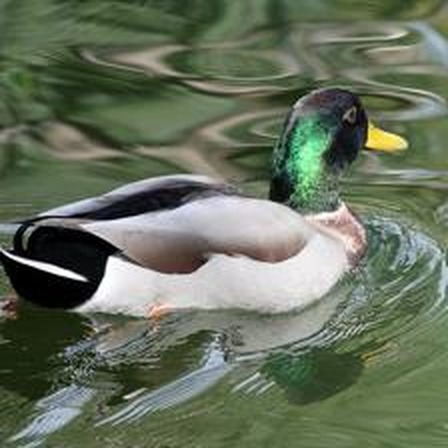}
\end{minipage}\hfill
\begin{minipage}[c]{0.56\textwidth}\small
\textbf{(a) MMVP.} Is the duck's entire beak visible in the picture?\\[3pt]
(a)~Yes \quad (b)~No\\[3pt]
\textbf{Ground truth:} (a)~Yes
\end{minipage}

\vspace{4pt}
{\small
\begin{tabular*}{\textwidth}{@{\extracolsep{\fill}}lccccccc}
\toprule
Model & Base & Vision-OPD & OPD-V & CV-OPD & VAD & OPD-Aha & \cellcolor{oursrow}\textbf{\method (ours)} \\
\midrule
Answer & (b)~\textcolor{BrickRed}{\xmark} & (b)~\textcolor{BrickRed}{\xmark} & (b)~\textcolor{BrickRed}{\xmark} & (b)~\textcolor{BrickRed}{\xmark} & (b)~\textcolor{BrickRed}{\xmark} & (b)~\textcolor{BrickRed}{\xmark} & \cellcolor{oursrow}\textbf{(a)}~\textcolor{ForestGreen}{\cmark} \\
\bottomrule
\end{tabular*}}

\vspace{10pt}
\noindent\begin{minipage}[c]{0.40\textwidth}\centering
\includegraphics[height=3.0cm, width=\linewidth, keepaspectratio]{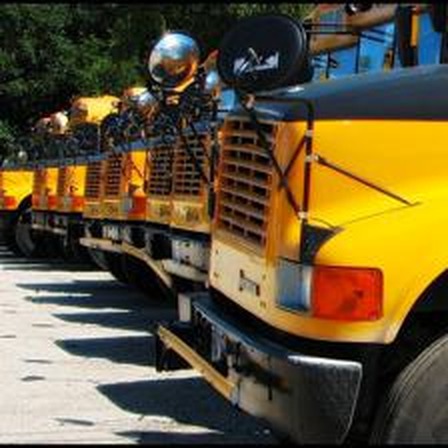}
\end{minipage}\hfill
\begin{minipage}[c]{0.56\textwidth}\small
\textbf{(b) MMVP.} Are the wheels of the school bus visible in the image?\\[3pt]
(a)~Yes \quad (b)~No\\[3pt]
\textbf{Ground truth:} (a)~Yes
\end{minipage}

\vspace{4pt}
{\small
\begin{tabular*}{\textwidth}{@{\extracolsep{\fill}}lccccccc}
\toprule
Model & Base & Vision-OPD & OPD-V & CV-OPD & VAD & OPD-Aha & \cellcolor{oursrow}\textbf{\method (ours)} \\
\midrule
Answer & (b)~\textcolor{BrickRed}{\xmark} & (b)~\textcolor{BrickRed}{\xmark} & (b)~\textcolor{BrickRed}{\xmark} & (b)~\textcolor{BrickRed}{\xmark} & (b)~\textcolor{BrickRed}{\xmark} & (b)~\textcolor{BrickRed}{\xmark} & \cellcolor{oursrow}\textbf{(a)}~\textcolor{ForestGreen}{\cmark} \\
\bottomrule
\end{tabular*}}

\vspace{10pt}
\noindent\begin{minipage}[c]{0.40\textwidth}\centering
\includegraphics[height=3.0cm, width=\linewidth, keepaspectratio]{figs/qual/gp_streetlight.jpg}
\end{minipage}\hfill
\begin{minipage}[c]{0.56\textwidth}\small
\textbf{(c) CV-Bench.} How many street lights are in the image?\\[3pt]
A.~2 \quad B.~3 \quad C.~0 \quad D.~1\\[3pt]
\textbf{Ground truth:} D.~1
\end{minipage}

\vspace{4pt}
{\small
\begin{tabular*}{\textwidth}{@{\extracolsep{\fill}}lccccccc}
\toprule
Model & Base & Vision-OPD & OPD-V & CV-OPD & VAD & OPD-Aha & \cellcolor{oursrow}\textbf{\method (ours)} \\
\midrule
Answer & A~\textcolor{BrickRed}{\xmark} & A~\textcolor{BrickRed}{\xmark} & A~\textcolor{BrickRed}{\xmark} & A~\textcolor{BrickRed}{\xmark} & A~\textcolor{BrickRed}{\xmark} & A~\textcolor{BrickRed}{\xmark} & \cellcolor{oursrow}\textbf{D}~\textcolor{ForestGreen}{\cmark} \\
\bottomrule
\end{tabular*}}

\vspace{10pt}
\noindent\begin{minipage}[c]{0.40\textwidth}\centering
\includegraphics[height=3.0cm, width=\linewidth, keepaspectratio]{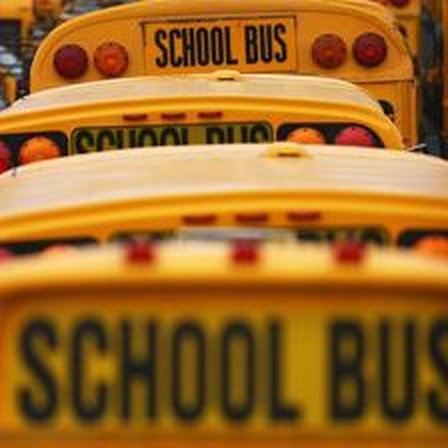}
\end{minipage}\hfill
\begin{minipage}[c]{0.56\textwidth}\small
\textbf{(d) MMVP.} From the camera's perspective, are the words ``SCHOOL BUS'' in the picture facing straight or oriented at an angle?\\[3pt]
(a)~Directly towards the camera\\[3pt]
(b)~At an angle\\[3pt]
\textbf{Ground truth:} (a)~Directly towards the camera
\end{minipage}

\vspace{4pt}
{\small
\begin{tabular*}{\textwidth}{@{\extracolsep{\fill}}lccccccc}
\toprule
Model & Base & Vision-OPD & OPD-V & CV-OPD & VAD & OPD-Aha & \cellcolor{oursrow}\textbf{\method (ours)} \\
\midrule
Answer & (b)~\textcolor{BrickRed}{\xmark} & (b)~\textcolor{BrickRed}{\xmark} & (b)~\textcolor{BrickRed}{\xmark} & (b)~\textcolor{BrickRed}{\xmark} & (b)~\textcolor{BrickRed}{\xmark} & (b)~\textcolor{BrickRed}{\xmark} & \cellcolor{oursrow}\textbf{(a)}~\textcolor{ForestGreen}{\cmark} \\
\bottomrule
\end{tabular*}}

\caption{Qualitative examples of general perception. Left: the test image. Right: the question, its options and the ground truth. Bottom: the answer of each model (\textcolor{ForestGreen}{\cmark}~correct, \textcolor{BrickRed}{\xmark}~wrong).}
\label{fig:qual-perc}
\end{figure*}

\begin{figure*}[p]
\centering
\noindent\begin{minipage}[c]{0.40\textwidth}\centering
\includegraphics[height=3.0cm, width=\linewidth, keepaspectratio]{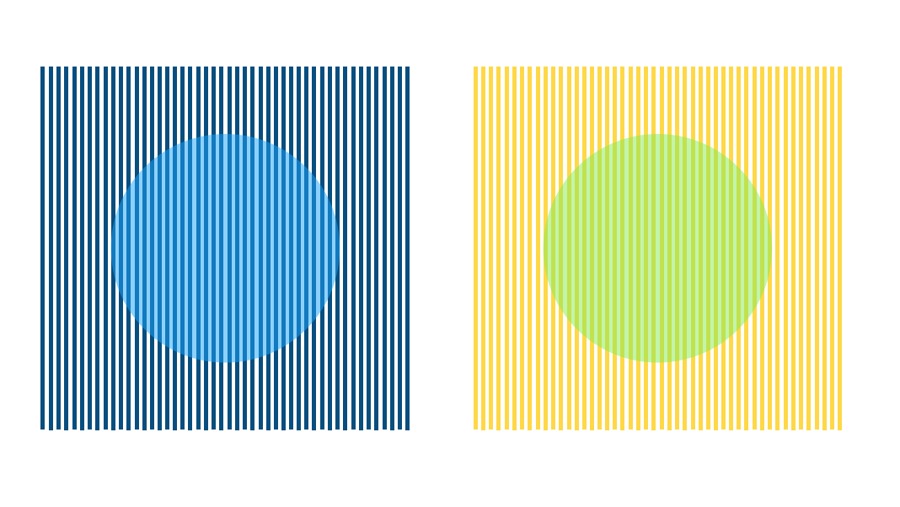}
\end{minipage}\hfill
\begin{minipage}[c]{0.56\textwidth}\small
\textbf{(a) HallusionBench.} The image shows a color illusion. Are two circles in the image different color?\\[3pt]
\textbf{Ground truth:} Yes
\end{minipage}

\vspace{4pt}
{\small
\begin{tabular*}{\textwidth}{@{\extracolsep{\fill}}lccccccc}
\toprule
Model & Base & Vision-OPD & OPD-V & CV-OPD & VAD & OPD-Aha & \cellcolor{oursrow}\textbf{\method (ours)} \\
\midrule
Answer & No~\textcolor{BrickRed}{\xmark} & No~\textcolor{BrickRed}{\xmark} & No~\textcolor{BrickRed}{\xmark} & No~\textcolor{BrickRed}{\xmark} & No~\textcolor{BrickRed}{\xmark} & No~\textcolor{BrickRed}{\xmark} & \cellcolor{oursrow}\textbf{Yes}~\textcolor{ForestGreen}{\cmark} \\
\bottomrule
\end{tabular*}}

\vspace{10pt}
\noindent\begin{minipage}[c]{0.40\textwidth}\centering
\includegraphics[height=3.0cm, width=\linewidth, keepaspectratio]{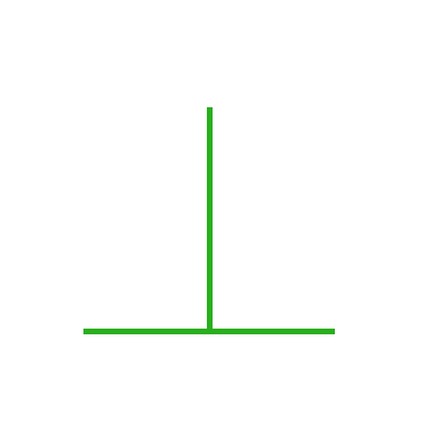}
\end{minipage}\hfill
\begin{minipage}[c]{0.56\textwidth}\small
\textbf{(b) HallusionBench.} Are the lengths of the vertical line and the horizontal line the same?\\[3pt]
\textbf{Ground truth:} Yes
\end{minipage}

\vspace{4pt}
{\small
\begin{tabular*}{\textwidth}{@{\extracolsep{\fill}}lccccccc}
\toprule
Model & Base & Vision-OPD & OPD-V & CV-OPD & VAD & OPD-Aha & \cellcolor{oursrow}\textbf{\method (ours)} \\
\midrule
Answer & No~\textcolor{BrickRed}{\xmark} & No~\textcolor{BrickRed}{\xmark} & No~\textcolor{BrickRed}{\xmark} & No~\textcolor{BrickRed}{\xmark} & No~\textcolor{BrickRed}{\xmark} & No~\textcolor{BrickRed}{\xmark} & \cellcolor{oursrow}\textbf{Yes}~\textcolor{ForestGreen}{\cmark} \\
\bottomrule
\end{tabular*}}

\vspace{10pt}
\noindent\begin{minipage}[c]{0.40\textwidth}\centering
\includegraphics[height=3.0cm, width=\linewidth, keepaspectratio]{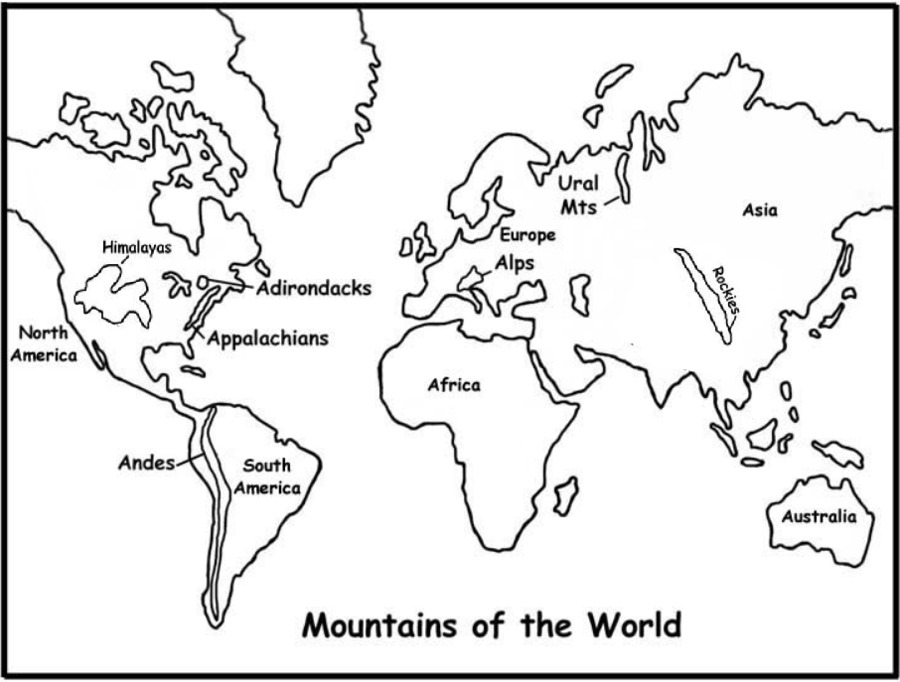}
\end{minipage}\hfill
\begin{minipage}[c]{0.56\textwidth}\small
\textbf{(c) HallusionBench.} Based on the map, are the Rocky Mountains located in North America?\\[3pt]
\textbf{Ground truth:} No
\end{minipage}

\vspace{4pt}
{\small
\begin{tabular*}{\textwidth}{@{\extracolsep{\fill}}lccccccc}
\toprule
Model & Base & Vision-OPD & OPD-V & CV-OPD & VAD & OPD-Aha & \cellcolor{oursrow}\textbf{\method (ours)} \\
\midrule
Answer & Yes~\textcolor{BrickRed}{\xmark} & Yes~\textcolor{BrickRed}{\xmark} & Yes~\textcolor{BrickRed}{\xmark} & Yes~\textcolor{BrickRed}{\xmark} & No~\textcolor{ForestGreen}{\cmark} & Yes~\textcolor{BrickRed}{\xmark} & \cellcolor{oursrow}\textbf{No}~\textcolor{ForestGreen}{\cmark} \\
\bottomrule
\end{tabular*}}

\vspace{10pt}
\noindent\begin{minipage}[c]{0.40\textwidth}\centering
\includegraphics[height=3.0cm, width=\linewidth, keepaspectratio]{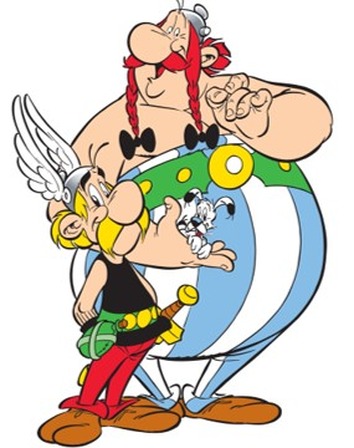}
\end{minipage}\hfill
\begin{minipage}[c]{0.56\textwidth}\small
\textbf{(d) HallusionBench.} Is there a Snoopy in this figure?\\[3pt]
\textbf{Ground truth:} No
\end{minipage}

\vspace{4pt}
{\small
\begin{tabular*}{\textwidth}{@{\extracolsep{\fill}}lccccccc}
\toprule
Model & Base & Vision-OPD & OPD-V & CV-OPD & VAD & OPD-Aha & \cellcolor{oursrow}\textbf{\method (ours)} \\
\midrule
Answer & No~\textcolor{ForestGreen}{\cmark} & Yes~\textcolor{BrickRed}{\xmark} & Yes~\textcolor{BrickRed}{\xmark} & Yes~\textcolor{BrickRed}{\xmark} & Yes~\textcolor{BrickRed}{\xmark} & Yes~\textcolor{BrickRed}{\xmark} & \cellcolor{oursrow}\textbf{No}~\textcolor{ForestGreen}{\cmark} \\
\bottomrule
\end{tabular*}}

\caption{Qualitative examples of hallucination. The layout is the same as in \cref{fig:qual-perc}.}
\label{fig:qual-hall}
\end{figure*}

\begin{figure*}[p]
\centering
\noindent\begin{minipage}[c]{0.40\textwidth}\centering
\includegraphics[height=3.0cm, width=\linewidth, keepaspectratio]{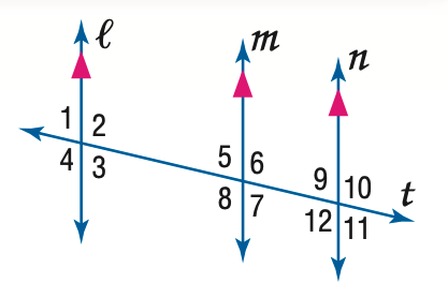}
\end{minipage}\hfill
\begin{minipage}[c]{0.56\textwidth}\small
\textbf{(a) MathVista.} In the figure, the measure of angle 9 is $75^\circ$. Find the measure of angle 6.\\[3pt]
A.~75 \quad B.~85 \quad C.~95 \quad D.~105\\[3pt]
\textbf{Ground truth:} D.~105
\end{minipage}

\vspace{4pt}
{\small
\begin{tabular*}{\textwidth}{@{\extracolsep{\fill}}lccccccc}
\toprule
Model & Base & Vision-OPD & OPD-V & CV-OPD & VAD & OPD-Aha & \cellcolor{oursrow}\textbf{\method (ours)} \\
\midrule
Answer & A~\textcolor{BrickRed}{\xmark} & A~\textcolor{BrickRed}{\xmark} & A~\textcolor{BrickRed}{\xmark} & A~\textcolor{BrickRed}{\xmark} & A~\textcolor{BrickRed}{\xmark} & A~\textcolor{BrickRed}{\xmark} & \cellcolor{oursrow}\textbf{D}~\textcolor{ForestGreen}{\cmark} \\
\bottomrule
\end{tabular*}}

\vspace{10pt}
\noindent\begin{minipage}[c]{0.40\textwidth}\centering
\includegraphics[height=3.0cm, width=\linewidth, keepaspectratio]{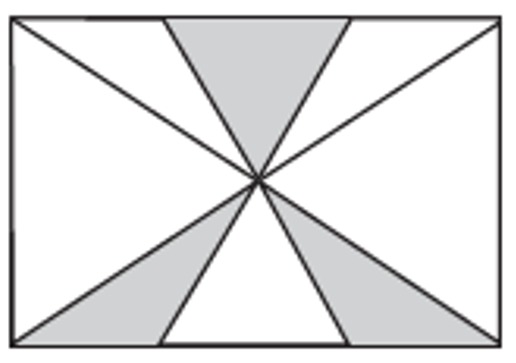}
\end{minipage}\hfill
\begin{minipage}[c]{0.56\textwidth}\small
\textbf{(b) MathVerse.} What fraction of the area of the figure is shaded?\\[3pt]
A.~3/8 \quad B.~1/4 \quad C.~1/8 \quad D.~1/10 \quad E.~1/16\\[3pt]
\textbf{Ground truth:} B.~1/4
\end{minipage}

\vspace{4pt}
{\small
\begin{tabular*}{\textwidth}{@{\extracolsep{\fill}}lccccccc}
\toprule
Model & Base & Vision-OPD & OPD-V & CV-OPD & VAD & OPD-Aha & \cellcolor{oursrow}\textbf{\method (ours)} \\
\midrule
Answer & A~\textcolor{BrickRed}{\xmark} & A~\textcolor{BrickRed}{\xmark} & A~\textcolor{BrickRed}{\xmark} & A~\textcolor{BrickRed}{\xmark} & A~\textcolor{BrickRed}{\xmark} & A~\textcolor{BrickRed}{\xmark} & \cellcolor{oursrow}\textbf{B}~\textcolor{ForestGreen}{\cmark} \\
\bottomrule
\end{tabular*}}

\vspace{10pt}
\noindent\begin{minipage}[c]{0.40\textwidth}\centering
\includegraphics[height=3.0cm, width=\linewidth, keepaspectratio]{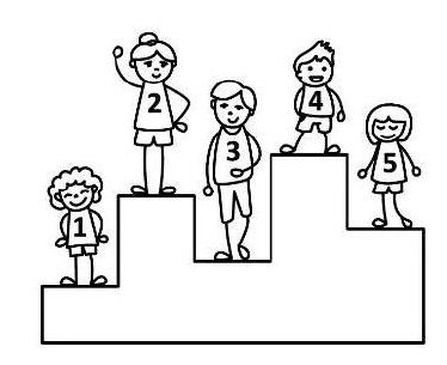}
\end{minipage}\hfill
\begin{minipage}[c]{0.56\textwidth}\small
\textbf{(c) MATH-Vision.} The higher someone stands on the podium, the better the ranking. Which number got third place?\\[3pt]
\textbf{Ground truth:} 5
\end{minipage}

\vspace{4pt}
{\small
\begin{tabular*}{\textwidth}{@{\extracolsep{\fill}}lccccccc}
\toprule
Model & Base & Vision-OPD & OPD-V & CV-OPD & VAD & OPD-Aha & \cellcolor{oursrow}\textbf{\method (ours)} \\
\midrule
Answer & 4~\textcolor{BrickRed}{\xmark} & 3~\textcolor{BrickRed}{\xmark} & 3~\textcolor{BrickRed}{\xmark} & 3~\textcolor{BrickRed}{\xmark} & 3~\textcolor{BrickRed}{\xmark} & 3~\textcolor{BrickRed}{\xmark} & \cellcolor{oursrow}\textbf{5}~\textcolor{ForestGreen}{\cmark} \\
\bottomrule
\end{tabular*}}

\vspace{10pt}
\noindent\begin{minipage}[c]{0.40\textwidth}\centering
\includegraphics[height=3.0cm, width=\linewidth, keepaspectratio]{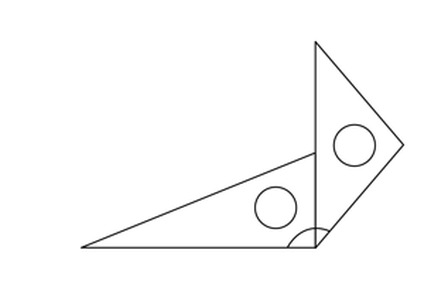}
\end{minipage}\hfill
\begin{minipage}[c]{0.56\textwidth}\small
\textbf{(d) We-Math.} As shown in the diagram, Xiaoxue used a set of triangular plates to piece together an angle. Its measure is (\quad).\\[3pt]
A.~$105^\circ$ \quad B.~$135^\circ$\\[3pt]
C.~$120^\circ$ \quad D.~No correct answer\\[3pt]
\textbf{Ground truth:} B.~$135^\circ$
\end{minipage}

\vspace{4pt}
{\small
\begin{tabular*}{\textwidth}{@{\extracolsep{\fill}}lccccccc}
\toprule
Model & Base & Vision-OPD & OPD-V & CV-OPD & VAD & OPD-Aha & \cellcolor{oursrow}\textbf{\method (ours)} \\
\midrule
Answer & A~\textcolor{BrickRed}{\xmark} & A~\textcolor{BrickRed}{\xmark} & A~\textcolor{BrickRed}{\xmark} & A~\textcolor{BrickRed}{\xmark} & A~\textcolor{BrickRed}{\xmark} & A~\textcolor{BrickRed}{\xmark} & \cellcolor{oursrow}\textbf{B}~\textcolor{ForestGreen}{\cmark} \\
\bottomrule
\end{tabular*}}

\caption{Qualitative examples of multimodal reasoning. The layout is the same as in \cref{fig:qual-perc}.}
\label{fig:qual-reas}
\end{figure*}

\noindent We show qualitative examples for the four abilities evaluated in \cref{tab:main,tab:zeroshot}: fine-grained perception (\cref{fig:qual}), general perception (\cref{fig:qual-perc}), hallucination (\cref{fig:qual-hall}) and multimodal reasoning (\cref{fig:qual-reas}).

\FloatBarrier

{
    \small
    \bibliographystylesupp{ieeenat_fullname}
    \bibliographysupp{main}
}

%% file: sec/2_related.tex
\noindent\textbf{\Large \textcolor{red}{L}. Related work}\\

\noindent\textbf{From thinking with images to a single-pass recipe for fine-grained multimodal perception.}
Inspired by ``thinking with images''~\cite{o3thinking2025}, a growing body of work improves fine-grained perception by progressively inspecting visual details through cropping and zooming at inference time. These approaches range from training-free visual search~\citesupp{vicrop2025,zoomeye2024} to reinforcement learning-based methods that learn to invoke visual tools~[\citealp{deepeyes2025,deepeyesv2_2026,thyme2025,sensenovamars2026}, \citealpsupp{minio32025,pixelreasoner2025}]. However, these methods often incur substantial computational overhead and latency during inference, limiting their applicability in real-time applications. A complementary line therefore seeks to amortize these capabilities into the model itself during post-training, enabling fine-grained visual details to be recognized in a single forward pass. Representative approaches include enhancing visual representations~\cite{chen2025shallower,barrios2026moda}, region-to-image distillation~\cite{zwz2026}, and OPD~\cite{visionopd2026,aha2026,vad2026,ead2026,vzero2026,rpopsd2026,vicur2026}. We follow this single-pass line: \method changes only the training target and leaves inference untouched.

\noindent\textbf{OPSD for fine-grained multimodal perception.}
OPD can improve fine-grained perception by distilling knowledge from a stronger teacher~\cite{vzero2026}. However, stronger teachers are not always available, particularly for frontier models. OPSD instead leverages privileged visual information to enable self-distillation. Representative methods equip the teacher with zoomed crops~\cite{visionopd2026}, higher-resolution images~\cite{rpopsd2026}, or recoverable visual cues~\cite{vicur2026}. Following the line of OPSD methods that use zoomed crops as privileged visual inputs, \method focuses on reconstructing distillation targets to counteract linguistic momentum and strengthen corrective supervision.

\noindent\textbf{Linguistic momentum and contrastive targets in multimodal OPSD.}
In OPD, conditioning the teacher on student-generated prefixes can bias its predictions toward the student's own mistakes, weakening corrective supervision~\citesupp{huang2026offpolicyteacher,gan2026raopd,akhondzadeh2026rgopd}. In multimodal OPSD, OPD-Aha~\cite{aha2026} further demonstrates that such linguistic momentum can override privileged visual evidence. To mitigate this issue, contrastive methods reconstruct the distillation target by contrasting the privileged teacher's prediction with a reference prediction conditioned on the same student prefix but with altered or removed visual input. Existing references include gray images~\cite{aha2026}, black images~\cite{vcsd2026,opdv2026}, Gaussian-noised images~\cite{leng2024vcd}, mismatched images~\cite{cvopd2026}, and text-only inputs~\cite{decomposedopd2026}. However, these references can leave residual linguistic drift and introduce false visual beliefs. In contrast, \method removes visual-content tokens while preserving the original multimodal input structure, enabling more effective cancellation of linguistic momentum and stronger corrective supervision.